\documentclass[10pt,letterpaper]{article}
\usepackage[top=0.85in,left=0.85in,right=0.85in,footskip=0.75in]{geometry}
\usepackage{amsmath,amssymb}
\usepackage{changepage}
\usepackage{textcomp,marvosym}
\usepackage{float}
\usepackage{cite}
\usepackage{nameref,hyperref}

\usepackage[table]{xcolor}

\usepackage{array}
\newcolumntype{+}{!{\vrule width 2pt}}

\newlength\savedwidth

\makeatletter
\renewcommand{\@biblabel}[1]{\quad#1.}
\makeatother

\usepackage{lastpage,fancyhdr,graphicx}
\usepackage{epstopdf}
\newcommand{\specialcell}[2][c]{%
  \begin{tabular}[#1]{@{}c@{}}#2\end{tabular}}

\usepackage{lastpage} 

\usepackage{fontspec}
\usepackage{polyglossia}
\setmainlanguage{english}
\setotherlanguage{bengali}
\newfontfamily\bengalifont{Kalpurush.ttf}[Path=./, Script=Bengali]
\newcommand{\bn}[1]{{\setmainfont{Kalpurush.ttf}[Path=./, Script=Bengali] #1}}

\usepackage[aboveskip=1pt,labelfont=bf,labelsep=period,justification=raggedright,singlelinecheck=off]{caption}
\addto\captionsenglish{%
  
}

\usepackage{booktabs}
\usepackage{multirow}

\usepackage{makecell}
\usepackage{rotating}
\usepackage{appendix}

\begin{document}
\vspace*{0.2in}

\begin{flushleft}
{\Large
\textbf{GHTM: A graph-based hybrid topic modeling approach for Bengali language}
}
\newline

Farhana Haque\textsuperscript{1},
Md. Abdur Rahman\textsuperscript{2,*},
Sumon Ahmed\textsuperscript{1,*},

\bigskip
\textbf{1} Institute of Information Technology (IIT), University of Dhaka, Dhaka, Bangladesh
\\
\textbf{2} Centre for Advanced Research in Sciences (CARS), University of Dhaka, Dhaka, Bangladesh
\\

\bigskip

*mukul.arahman@gmail.com (AR), sumon@du.ac.bd (SA)

\end{flushleft}

\section*{Abstract}
Topic modeling is a natural language processing technique used to discover latent themes and abstract topics from text corpora by grouping co-occurring keywords. Although widely researched in English, topic modeling remains understudied in Bengali due to a lack of adequate resources and initiatives. Existing Bengali topic modeling research lacks exploration of modern methodological approaches, reproducible implementations, diverse datasets, and systematic benchmarking. To bridge these gaps, this study presents Graph-based Hybrid Topic Model (GHTM), a novel architecture that synthesizes the distinct strengths of multiple paradigms by unifying statistical (TF-IDF), semantic (GloVe), and structural (GCN) modalities to construct semantically aligned document representations. GHTM later employs matrix factorization for topic extraction, offering a computationally lightweight alternative to resource-intensive neural and graph-based approaches. Through comprehensive benchmarking across three Bengali datasets of varying scales, GHTM consistently surpasses both traditional and contemporary topic modeling approaches, achieving improved topic coherence (NPMI: $0.27–0.28$) and high topic diversity while maintaining computational efficiency. The model also outperformed established graph-based models on the English \textit{20Newsgroups} benchmark. Additionally, we introduce \textit{NCTBText}, a diverse Bengali textbook-based dataset comprising 8,650 text documents, providing much-needed topical diversity beyond newspaper-centric Bengali corpora.

\section*{Introduction}

Topic modeling is a powerful, unsupervised text-mining technique in Natural Language Processing (NLP) that discovers latent semantic patterns in unstructured text corpora. In the era of exponential growth in digital text, the ability to automatically extract meaningful patterns from large document collections has become increasingly crucial for supporting diverse NLP tasks. Topic modeling addresses this challenge by creating insightful thematic word sets that reveal underlying topics. It makes sense of chaotic real-world data without the need for manual evaluation of large datasets. The discovered topics can further contribute to downstream NLP applications\cite{pergola2019tdam,yu2022semanticsts,SRIVASTAVA2022108636}, such as document classification, information retrieval, text summarization, and exploratory data analysis.

Topic modeling offers particularly compelling advantages for low-resource languages like Bengali by enabling the discovery of thematic structures from large volumes of text without reliance on costly expert-labeled datasets. Effective topic modeling can serve as a foundational component for a wide range of Bengali NLP tasks\cite{raihan2024overview,rahman2025topicmap}, strengthening the overall Bengali NLP infrastructure. Topic modeling has advanced remarkably since its emergence, progressing from foundational probabilistic models, such as Latent Dirichlet Allocation (LDA)\cite{blei2003latent}, and algebraic approaches, like Non-negative Matrix Factorization (NMF)\cite{lee1999learning}, to sophisticated neural architectures incorporating contextualized embeddings\cite{srivastava2017autoencoding,dieng2020topic,bianchi-etal-2021-pre,bianchi-etal-2021-cross}. Most recently, cluster-based\cite{angelov2020top2vec,grootendorst2022bertopic}, graph-based\cite{zhu2018graph,luo2023graph,adhya2025ginopic,zhang2023graph2topic,shen2021topic}, and Large Language Model (LLM)-powered approaches\cite{pham2023topicgpt} have further expanded the field. While the English NLP community has extensively explored these advanced techniques and established robust evaluation frameworks, topic modeling research for Bengali remains substantially underdeveloped. Despite being the seventh-most-spoken language globally, with over 284 million speakers\cite {ethnologue_bengali2025}, Bengali topic modeling has received limited attention. 

To date, a systematic comparison of topic models for Bengali remains absent from the literature, and existing studies in this area have largely been confined to LDA\cite{helal2018topic, hasan2019lda2vec, alam2020bengali, paul2025combining}. Furthermore, there is no widely accepted, standardized dataset specifically designed to evaluate Bengali topic models. Most available Bengali datasets are collected from online newspapers\cite{ahmad2022potrika,akash2023shironaam,saad2024bnad,hossain2020banfakenews}, with limited topical diversity. Together, these gaps have significantly impeded the development of Bengali topic modeling as a field.

This study addresses these critical research gaps by investigating the following research questions: (1) Which topic models, from traditional to contemporary, perform best for Bengali? (2) Is it possible for a novel model to outperform current models by combining top-performing architectures? (3) How does the new model perform on English benchmark datasets? (4) How do the novel model and existing topic modeling approaches perform on a new Bengali corpus curated from non-newspaper sources?

In pursuit of substantial answers to these research questions, we develop the \textbf{Graph-based Hybrid Topic Model (GHTM)}, a novel architecture that strategically integrates multiple complementary techniques to achieve superior topic quality. GHTM initially represents documents as TF-IDF-weighted GloVe embeddings\cite{pennington-etal-2014-glove}, to preserve domain-specific word importance alongside corpus-level semantic relationships. Then it constructs a k-nearest-neighbor graph where these document representations serve as nodes and edges represent weighted connections between semantically related documents. These representations are subsequently refined through Graph Convolutional Networks (GCN)\cite{kipf2017semi}, which smooth document-level noise and pull semantically related documents into compact, highly separable clusters. Finally, NMF is applied to the refined embeddings to extract interpretable topics, leveraging its deterministic interpretability and soft-assignment capabilities. 

GHTM differs from existing graph-based topic models across several paradigms. First, current approaches~\cite{shen2021topic, luo2023graph, adhya2025ginopic} rely on Variational Autoencoders (VAEs) and Graph Neural Networks (GNNs) to learn probabilistic, continuous latent topic distributions. In contrast, GHTM takes a deterministic approach to document-topic assignment, by feeding GNN-refined document representations to NMF. Second, rather than limiting convolutions to word co-occurrence biterms~\cite{zhu2018graph}, which capture only micro-level patterns, GHTM constructs a document-level similarity graph to preserve macro-level community topology. Third, unlike direct embedding clustering~\cite{zhang2023graph2topic}, which derives topic assignments through community detection, GHTM utilizes factorization to yield sharper fine-grained topic distributions. Another closely related work~\cite{dahogNMF} incorporates higher-order graph Laplacian regularization directly into the NMF loss function, preserving local document and word geometry during the decomposition process. GHTM, on the other hand, uses GCNs to reshape and encode document manifold topology before factorization. By separating graph representation learning from NMF, GHTM maintains geometric structure while keeping the downstream factorization objective straightforward and efficient.

While designed for Bengali, GHTM's architecture is language-agnostic and readily applicable to other languages. To validate its effectiveness beyond Bengali, we evaluate it alongside modern graph-based baseline models on the widely recognized English benchmark dataset \textit{20Newsgroups}. Also, to address the absence of systematic comparative studies in this field, we conduct a comprehensive evaluation of topic models across three Bengali datasets of varying characteristics and sizes. Two Bengali newspaper datasets are publicly available, while the third one is a novel Bengali textbook dataset that we curated from materials provided by Bangladesh’s \href{http://www.nctb.gov.bd}{National Curriculum and Textbook Board (NCTB)}. This new dataset, called \textbf{\textit{NCTBText}}, introduces much-needed topical diversity to the newspaper-dominated Bengali dataset landscape. Our experimental results demonstrate that GHTM consistently achieves superior topic coherence and diversity compared to both traditional and contemporary methods, establishing new performance benchmarks for Bengali topic modeling.

The major contributions of this study are as follows:
\begin{itemize}
  \item Comprehensive comparison of traditional and contemporary topic modeling methods across three Bengali datasets.
  \item Development of GHTM, a novel graph-based hybrid topic modeling approach.
  \item Performance evaluation of GHTM and modern graph-based models on the English benchmark dataset \textit{20Newsgroup}.
  \item Creation of \textit{NCTBText}, a diverse Bengali textbook-based dataset that expands beyond newspaper-centric corpora.
\end{itemize}

The paper is organized as follows. We begin by exploring the evolution of topic modeling paradigms and discussing the current state of Bengali topic modeling research, noting critical gaps that motivate this work. We then introduce our proposed model, GHTM, and the novel dataset, \textit{NCTBText}. Subsequently, we describe the experimental setup for the comparative analysis of topic models. Then, we report the results, accompanied by topic quality evaluation, ablation analysis, and statistical significance testing. Finally, we discuss the implications of our findings, acknowledge limitations, and conclude with directions for future research.

\section*{Related work}
This section provides a brief overview of the evolution of topic modeling architectures from its classical roots to contemporary approaches. Subsequently, we review the current landscape of topic modeling research in Bengali, highlighting its limited progress compared to mainstream studies and the key research gaps, that we aim to address in this work. 

\subsection*{Evolution of Topic Models}
Topic modeling methodologies have evolved substantially over the years, advancing from algebraic, probabilistic, and neural models to modern cluster-based, graph-centric, and LLM-driven paradigms. This subsection discusses the significant models that have shaped this field of study. 

\subsubsection*{Traditional models} 
Foundational topic modeling research was based on algebraic factorization and probabilistic techniques that relied on document-term, also known as Bag-of-Words (BoW) and later TF-IDF representations of text. Namely, Latent Semantic Analysis (\textbf{LSA}), also known as Latent Semantic Indexing (LSI)\cite{deerwester1990indexing}, is an early model that identifies latent semantic structures by applying Singular Value Decomposition (SVD) to document-term matrices. SVD reduces the sparsity of the matrices and compares the documents using cosine similarity in a lower-dimensional space. \textbf{NMF}\cite{lee1999learning}, on the other hand, factorizes a non-negative document-term matrix into two lower rank matrices. One represents document-topic distributions, while the other represents topic-word distributions. Another classic model is Latent Dirichlet Allocation (\textbf{LDA})\cite{blei2003latent}, which is widely popular and serves as a baseline for topic modeling across languages. It is a probabilistic generative model that identifies latent topics by inferring the probability distributions of topics, assuming that documents are mixtures of topics and that topics are mixtures of words. Although these models are cornerstones in topic modeling research and they have definitively paved the way for modern approaches, they share two fundamental constraints. Firstly, their sole reliance on BoW and TF-IDF for text representation fails to capture semantic relationships between words and limits their adaptability to leverage modern contextual embeddings. And secondly, as these models are CPU-bound, they cannot leverage the parallel processing capabilities of GPUs, which limits their scalability. 

\subsubsection*{Neural Topic Models (NTMs)} 
Building upon the probabilistic foundation of topic models, NTMs came into light as neural network architectures got popular, and GPU acceleration became accessible. These models embraced Neural Variational Inference (NVI) framework\cite{mnih2014neural}, which is inspired by the Variational Auto Encoder (VAE)\cite{kingma2013auto}. They employ continuous latent representations and deep neural networks to capture richer semantic structures. Notably, Autoencoding Variational Inference for Topic Models (AVITM) \cite{srivastava2017autoencoding} introduced \textbf{ProdLDA}, a neural topic model that employs a VAE to encode a BoW vector into a Gaussian latent topic representation, which is then decoded using a product of multinomial experts to reconstruct the document’s word distribution. Alternatively, Embedded Topic Model (\textbf{ETM})\cite{dieng2020topic} models topics as distributions over word embeddings by embedding words and topics in a shared semantic space. Drawing inspiration from these findings, \textbf{CombinedTM}\cite{bianchi-etal-2021-pre} enhances ProdLDA by concatenating BoW with Sentence-BERT (SBERT) embeddings as VAE input. Another approach from the same authors, \textbf{ZeroShotTM}\cite{bianchi-etal-2021-cross} replaces BoW in ProdLDA with SBERT embeddings for zero-shot topic transfer, leveraging semantic embeddings to infer topics without domain-specific training. 
%The study also implements \textbf{NeuralLDA}, which mimics the classical mixture-based formulation of LDA within the same variational framework, where each document’s word distribution is modeled as a weighted mixture of topic–word distributions.

\subsubsection*{Cluster-based models} 
In more recent years, the emergence of transformer-based architectures like BERT introduced high-quality contextual sentence embeddings. This phenomenon enabled a paradigm shift in this field by treating topic modeling as a document-clustering problem in high-dimensional semantic space, with each cluster representing a topic. For instance, \textbf{Top2Vec}\cite{angelov2020top2vec} jointly embeds documents and words, reduces dimension with UMAP and clusters them using density-based clustering (HDBSCAN) to detect topics. In a comparable manner, \textbf{BERTopic}\cite{grootendorst2022bertopic} utilizes BERT embeddings, UMAP, HDBSCAN, and class-based TF-IDF for topic-word extraction, to form a modular pipeline for topic modeling with a data-driven topic count.  \textbf{CluWords}\cite{viegas2019cluwords} is another influential approach that leverages pre-trained word embeddings to group semantically related words. For each word, the model identifies its k-nearest neighbors in the embedding space and forms clusters of words that capture distributional semantics. Documents are then represented as weighted combinations of these words, termed “CluWords”.  

\subsubsection*{Graph-based models} 
Graph Neural Networks (GNNs) have been proven to be effective across many scientific tasks~\cite{Zheng2026}, including NLP, and have recently piqued the interest of the topic modeling community. GNNs enabled topic models to capture structural relationships between words and documents, as well as complex semantic dependencies. More recent studies have utilized a variety of GNNs in topic modeling, which has advanced this field even further. Namely, \textbf{GraphBTM}\cite{zhu2018graph} represents bi-terms as graphs and designs GCNs with residual connections to extract transitive features from bi-terms, resulting in coherent topics. Another method called Graph Neural Topic Model (\textbf{GNTM}) \cite{shen2021topic} represents documents as semantic graphs and uses the NVI approach with GNN for topic modeling. \textbf{Graph2Topic}\cite{zhang2023graph2topic}, on the other hand, uses sentence embeddings and community detection algorithms on graph structures for topic modeling, framing topic discovery as a community finding problem in a semantic graph. Graph Contrastive Neural Topic Model (\textbf{GCTM})\cite{luo2023graph} is another approach that integrates contrastive learning with topic modeling via graph-based sampling. Their model treats an input document as a document word bipartite graph and constructs positive and negative word co-occurrence graphs to capture in-depth semantic correlation among words. Alternatively, \textbf{GINopic}\cite{adhya2025ginopic} utilizes word similarity graphs for each document, where the graph is constructed using word embeddings to capture the complex word-level co-dependencies. Then, to produce document embeddings from these graphs, the authors employ Graph Isomorphism Network (GIN) that captures the structural relationships among words. 

\subsubsection*{LLM-based models} 
The most recent breakthrough in topic modeling is LLM-based architectures, which leverage LLM's knowledge-based generative capabilities. Some research efforts\cite{pham2023topicgpt,mu2024large,mu2024addressing} explore the idea of zero-shot topic extraction based on LLM's inherent knowledge and prompting techniques, overcoming the need for training models. \textbf{TopicGPT}\cite{pham2023topicgpt} is a notable example of this category, as it presents a framework that uses LLMs to generate and refine topics that are more interpretable and human-aligned than those discovered by traditional topic modeling methodologies. 

\subsection*{Topic modeling in Bengali}
Despite significant advancements in topic modeling architectures over the years, topic modeling efforts in Bengali haven't evolved considerably. The field currently suffers from multiple critical deficiencies that collectively hinder meaningful progress. This subsection reviews key publications on Bengali topic modeling and identifies significant research gaps that we wish to address in this study.

The pioneering work by \cite{helal2018topic} took the first step towards Bengali topic modeling by applying LDA on a Bengali dataset, establishing a baseline for subsequent research. The study shows the efficacy of LDA with bigrams for Bengali news classification and topic extraction. Another subsequent study\cite{hasan2019lda2vec} shows LDA2Vec's high accuracy over LDA itself, in a comparison between these two models on a Bengali newspaper dataset. LDA2Vec\cite{moody2016mixing} is a hybrid approach that ties the interpretability of LDA with the semantic power of word2vec. An alternative study~\cite {alam2020bengali} curated a Bengali dataset consisting of 70K news articles and applied LDA to uncover latent topics and observe media trend evolution over time in Bengali news. The study offers an in-depth analysis and demonstrates how different topics prevail across weeks. A comprehensive survey\cite{ahmed2021systematic} examines the landscape of topic modeling research in Bengali from 2003 through 2020. It highlights the disparity between English and Bengali topic modeling efforts, answering some of the most important questions, such as ``What are the techniques that have been used in English topic modeling but not yet used in Bengali?", ``What are the sources of the datasets used?", etc., through rigorous research. The study also effectively outlines future research scopes for Bengali topic modeling.

Recent years have witnessed growing efforts to develop topic modeling approaches tailored specifically for Bengali. A notable contribution\cite{paul2025combining} compiled a novel Bengali news dataset and proposed a hybrid model combining the potentials of both LDA and BERT, called BERT-LDA, advancing topic modeling in Bengali. The study compares its proposed hybrid model with traditional models such as LDA, LSI, and the Hierarchical Dirichlet Process (HDP) in terms of topic coherence. The authors also applied their model to English benchmark datasets (\textit{20NewsGroup}, \textit{BBC}) and demonstrated the results. Another research effort\cite{dawn2024likelihood} in this area proposes a Dirichlet-polynomial clustering model called Likelihood Corpus Distribution (LCD), which is based on a Bayesian numerical prototype that evaluates the probability distribution of words in a document to identify topics. Experiments demonstrate the efficacy of LCD compared to traditional topic models across five extensive Bengali themes curated by the authors. More recently, an approach\cite{rifat2025clustering} proposes to cluster word vectors derived from BanglaBERT\cite{bbert2022}, a transformer-based language model pretrained on Bengali. Their method suggests TF-IDF-weighted clustering and reordering to identify topic words, which demonstrates superior performance over LDA on a Bengali dataset of 197,238 news articles, compiled by the authors.

Beyond standalone topic discovery, two studies have leveraged topic modeling as a foundational component for broader NLP applications in Bengali. One publication\cite{raihan2024overview} shows topic modeling's utility for exploratory data analysis of Bengali social media discourse. The study employed BERTopic to analyze speech trends in social media data as part of their hate speech identification framework. Similarly, another study\cite{rahman2025topicmap} introduces the TopicMap-BN framework, which integrates BERTopic with BanglaBERT for large-scale news recommendation systems. Their approach demonstrates that topic modeling can serve as an effective intermediate layer for recommendation systems that require interpretable content categorization.

The review of these Bengali topic modeling studies reveals several critical research gaps. To begin with, there is a definitive absence of widespread comparative studies evaluating both modern approaches and traditional methods, which represents a critical gap in understanding which architectural choices are most suitable for Bengali data. The absence of such systematic benchmarking prevents researchers from making informed architectural choices and objectively assessing incremental improvements.

Moreover, the studies employed various evaluation metrics, including perplexity\cite{dawn2024likelihood}, coherence scores ($C_V$\cite{dawn2024likelihood,paul2025combining}, NPMI\cite{rahman2025topicmap}, average NPMI\cite{rifat2025clustering}), cluster validity indices (silhouette score, calinski-harabasz index, davies-bouldin index)\cite{paul2025combining}, similarity measures (cosine similarity\cite{helal2018topic}, Jaccard similarity\cite{paul2025combining}), accuracy\cite{hasan2019lda2vec}, topic-label correspondence\cite{alam2020bengali} and topic diversity, with no consistent use of a specific set of metrics and evaluation protocol. This inconsistency renders cross-study comparisons infeasible and obscures relative model performance.

In addition, the majority of studies limit their exploration to either LDA\cite{helal2018topic,alam2020bengali} and its extensions (LDA2Vec\cite{hasan2019lda2vec}, BERT-LDA\cite{paul2025combining}), or more recently, cluster-based methods\cite{raihan2024overview,rifat2025clustering}, with no adaptation of modern approaches such as neural topic models, graph-based architectures, or LLM-based frameworks that have demonstrated superior performance in English. Consequently, the field has only three novel architectural contributions designed explicitly for Bengali: LCD\cite{dawn2024likelihood}, BERT-LDA\cite{paul2025combining}, and the BERT-based word embedding clustering approach\cite{rifat2025clustering}. Also, these studies mostly benchmark their results against classic models without considering contemporary baselines.

These limitations suggest that the Bengali topic modeling landscape remains significantly underexplored, particularly in terms of adapting state-of-the-art methodologies. Additionally, there is no ``gold standard" Bengali dataset like \textit{20NewsGroup} for English, as each study evaluates on different datasets, which makes it impossible to objectively rank models. Most research relies solely on news article-based datasets for evaluation, which fail to stress-test a model's resilience against linguistic diversity, such as social media, academic writing, fiction, etc. Furthermore, to the best of our knowledge, none of the studies appear to have publicly released their implementation, which hinders reproducibility for further validation and research in this field.

Therefore, to bridge these gaps, our work presents a comprehensive evaluation of topic models in Bengali, employing a broad range of baselines, diverse datasets, and a standard set of evaluation metrics. We also introduce a reproducible, novel topic model that incorporates modern techniques and a benchmark dataset to facilitate future comparative research in this field.

\section*{Graph-based Hybrid Topic Model (GHTM)}
This section introduces the Graph-based Hybrid Topic Model (GHTM), a novel framework for discovering latent topics and semantically coherent topic words from document collections. GHTM integrates statistical word weighting, semantic word embeddings, graph neural networks, and matrix factorization into a unified four-stage pipeline. The model: (1) converts text documents into hybrid vector representations, (2) constructs a similarity graph to define topical areas and refine document embeddings through graph convolutions, (3) decomposes the embeddings to identify latent topics from document-topic distributions, and (4) finally extracts interpretable keywords to represent the discovered topics. Fig~\ref{fig:Fig1} illustrates the overall architecture of the model, while the following subsections detail each stage. Table~\ref{table:notation} summarizes the mathematical notations used throughout this section.

\subsection*{Text vectorization}
The first stage transforms input text documents into dense vector representations that capture both the statistical importance and semantic meaning of their constituent words. Since our ultimate goal is to represent topics through keyword sets, we vectorize documents at the word level, then aggregate these word embeddings into document vectors. 

To vectorize the words that we get after tokenizing and cleaning the raw documents, we employ a dual-component vectorization technique. This hybrid technique combines statistical word weighting with contextual word embeddings~\cite{7259377,Seegmiller2023,Schmidt2019}. This approach addresses a fundamental limitation: sparse representations like bag-of-words or TF-IDF weighting only capture term importance but ignore semantic relationships, while word embedding models capture semantics but treat all words equally. By integrating both to vectorize the documents, we capture document-level term importance and corpus-level semantic relationships simultaneously. This provides a robust foundation for the subsequent stage to identify meaningful document communities that correspond to latent topics.

We first compute Term Frequency-Inverse Document Frequency (TF-IDF) weights to quantify word importance across the document collection. Term Frequency (TF) measures how often a word appears in a specific document, while Inverse Document Frequency (IDF) reflects how common or rare the word is across all documents. Words appearing in many documents are considered less informative and are accordingly downweighted. Given a corpus $\mathcal{D} = \{d_1, d_2, \ldots, d_N\}$ with vocabulary $\mathcal{V} = \{w_1, w_2, \ldots, w_V\}$, the TF-IDF weight for term $w_j$ in document $d_i$ is computed as:

\begin{equation}
\text{TF-IDF}(w_j, d_i) = \text{TF}(w_j, d_i) \times \text{IDF}(w_j)
\end{equation}

where the term frequency is defined as:

\begin{equation}
\text{TF}(w_j, d_i) = \frac{\text{count of } w_j \text{ in } d_i}{\text{total words in } d_i}
\end{equation}

and the inverse document frequency is defined as:

\begin{equation}
  \text{IDF}(w_j) = \log\left(\frac{N}{1 + \text{number of documents in $\mathcal{D}$ containing $w_j$}}\right)
\end{equation}

Here, $N$ represents the total number of documents. This produces a sparse document-term matrix $\mathbf{S} \in \mathbb{R}^{N \times V}$.

To complement TF-IDF's statistical weighting with semantic context, we employ pretrained GloVe embeddings, which capture fine-grained semantic relationships through global word co-occurrence patterns. Let $\mathbf{G} \in \mathbb{R}^{V \times D}$ represent the GloVe embedding matrix, where row $\mathbf{g}_j \in \mathbb{R}^D$ corresponds to the $D$-dimensional embedding vector for word $w_j$ from vocabulary $\mathcal{V}$. 

\begin{figure}[H]
  \includegraphics[width=\linewidth]{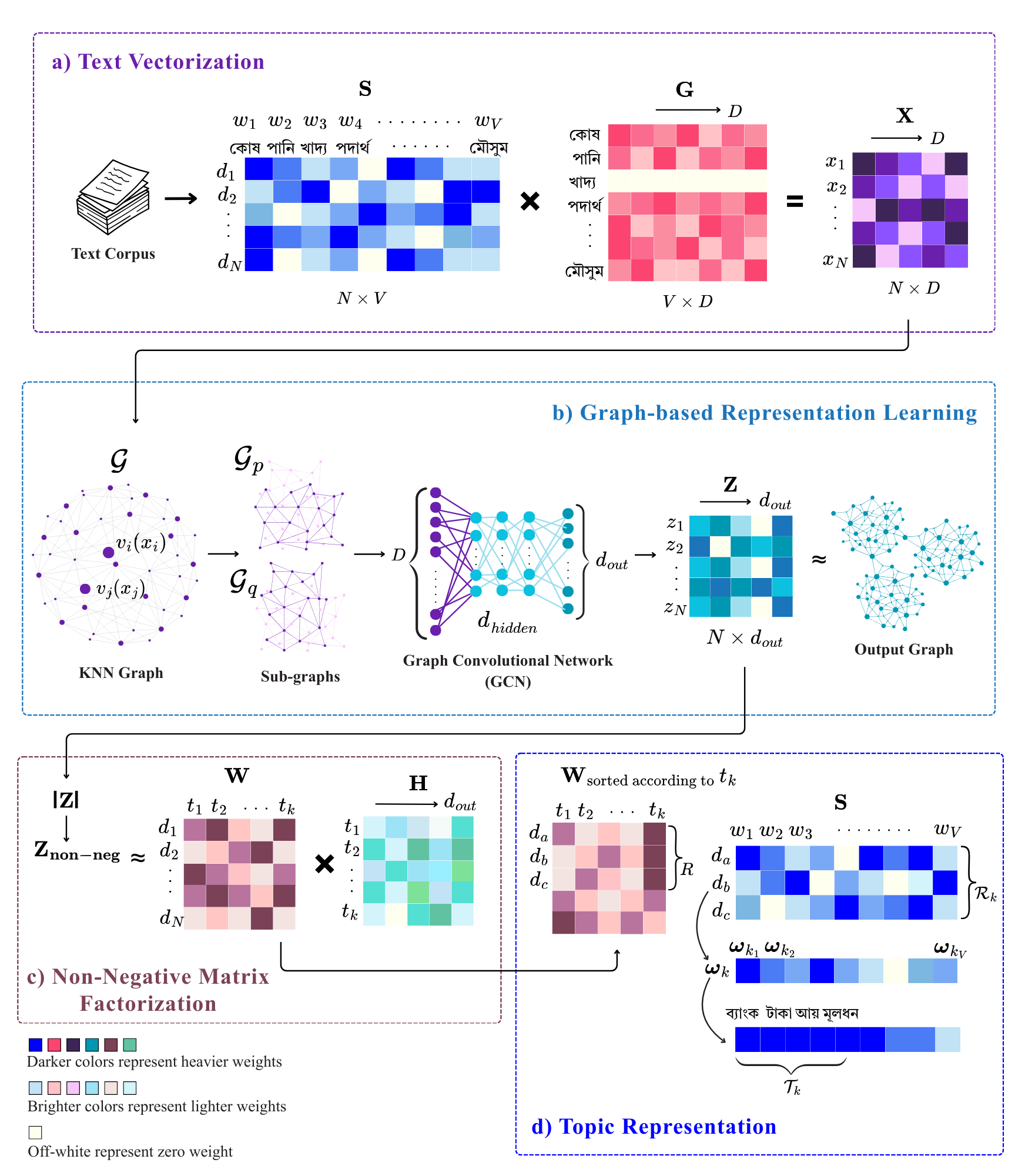}
  \caption{\textbf{Overview Diagram of Graph Hybrid Topic Model (GHTM) architecture.}
  a) Text Vectorization: This stage converts the text documents into a sum of TF-IDF weighted GloVe embeddings. b) Graph-based Representation Learning: This stage generates a KNN graph, and separates it into sub-graphs. Document vectors, represented as nodes in the graph, are enriched by the GCN layers, resulting in the output matrix $\mathbf{Z}$. The output graph, as shown in the diagram, densely clusters related nodes, which are now communally informed. c) Non-negative Matrix Factorization: This stage accomplishes topic modeling by first performing an absolute value transformation on $\mathbf{Z}$ and then factorizing it into $\mathbf{W}$ and $\mathbf{H}$. d) Topic Representation: Finally, this stage identifies the top $R$ representative documents from $\mathbf{W}$ and chooses a collection of keywords to represent the topics.}
\label{fig:Fig1}
  \end{figure}

The final document representation matrix is obtained through the matrix multiplication of $\mathbf{S}$ and $\mathbf{G}$:

\begin{equation}
\mathbf{X} = \mathbf{S} \cdot \mathbf{G}
\label{eq:multiply}
\end{equation}

where $\mathbf{X} \in \mathbb{R}^{N \times D}$ represents the dense document embedding matrix. Each document vector $\mathbf{x}_i$ is effectively a TF-IDF weighted sum of its constituent word embeddings:

\begin{equation}
  \mathbf{x}_i = \sum_{j=1}^{V} \text{TF-IDF}_{i,j} \cdot \mathbf{g}_j
  \label{eq:vector}
\end{equation}

This formulation ensures that semantically meaningful words are weighted by their document-specific importance. The resulting hybrid representations preserve both term relevance and semantic relationships, enabling effective document similarity computation for subsequent graph construction and topic discovery. Note that words with zero TF-IDF weights or missing GloVe embeddings (depicted as off-white cells in Fig~\ref{fig:Fig1}) contribute nothing to the final document vectors, as they are nullified during matrix multiplication.

\begin{table}[!ht]
  \centering
  \scriptsize
  \caption{\textbf{Notation summary for GHTM.}}
  \renewcommand{\arraystretch}{1.5}
  \begin{tabular}{|l|l|}
  \hline
  \textbf{Notation} & \textbf{Description} \\ \hline
  $\mathcal{D}$ & Corpus of documents, $\mathcal{D} = \{d_1, d_2, \ldots, d_N\}$ \\ \hline
  $N$ & Total number of documents in the corpus \\ \hline

  $V$ & Total number of unique terms in the vocabulary \\ \hline

  $\mathbf{S} \in \mathbb{R}^{N \times V}$ & Sparse document-term matrix with TF-IDF weights \\ \hline
  $\mathbf{G} \in \mathbb{R}^{V \times D}$ & GloVe embedding matrix for vocabulary \\ \hline

  $D$ & Dimensionality of GloVe word embeddings \\ \hline
  $\mathbf{X} \in \mathbb{R}^{N \times D}$ & Dense document embedding matrix (TF-IDF weighted GloVe) \\ \hline

  $\mathcal{G} = (\mathcal{V}, \mathcal{E})$ & K-Nearest Neighbor graph with nodes $\mathcal{V}$ and edges $\mathcal{E}$ \\ \hline
  $\mathcal{K}$ & Number of nearest neighbors in KNN graph construction \\ \hline

  $P$ & Number of clusters in Cluster-GCN partitioning \\ \hline

  $d_{\text{hidden}}$ & Dimensionality of hidden layers in GCN \\ \hline
  $d_{\text{out}}$ & Output dimensionality of final GCN layer \\ \hline
 
  $\mathbf{Z} \in \mathbb{R}^{N \times d_{\text{out}}}$ & Output embeddings from GCN \\ \hline
  
  $K$ & Number of topics \\ \hline
  $\mathbf{W} \in \mathbb{R}_+^{N \times K}$ & Document-topic distribution matrix from NMF \\ \hline
 
  $R$ & Number of representative documents per topic \\ \hline
  $\mathcal{R}_k$ & Set of top-$R$ representative documents for topic $k$ \\ \hline
  
  $T$ & Number of top keywords extracted per topic \\ \hline
  $\mathcal{T}_k$ & Set of top-$T$ keywords representing topic $k$ \\ \hline
  \end{tabular}
  \begin{flushleft} 
  %This table summarizes the mathematical notations used throughout GHTM.
  \end{flushleft}
  \label{table:notation}

\end{table}

\subsection*{Graph-based representation learning}

Having obtained document vectors in the previous stage, we now leverage their geometric relationships to discover latent topical structure. This intermediate stage transforms the initial hybrid vector representations into structurally informed dense embeddings through graph-based representation learning. This technique operates on the graph topology of the text corpus, which portrays relationships among the documents. It organizes inherent but vague communities in the embedding space into compact and coherent clusters based on topical similarity. Simultaneously, vector representations learn to reflect this improved community structure and are therefore transformed into structurally informed embeddings. This stage facilitates cleaner topic separation in the subsequent stage by defining distinct topical areas in the embedding space and educating the document embeddings about them.

We begin by constructing a document-similarity graph, treating each document vector as a node to capture the relational structure among documents in the embedding space. We then apply graph-based learning on the resulting graph to transform the document vectors so that topically related documents form densely connected communities and reflect this updated relational structure. As a result, the modified embeddings create an embedding space with well-defined cluster boundaries. In this section, we first describe the graph construction process and then detail the graph-based learning procedure that transforms the vector representations into community-structured document embeddings.

\subsubsection*{Graph construction} To represent the document vectors as a graph, document similarity relationships are modeled through a K-Nearest Neighbour (KNN) graph. Document embeddings in the vector space are considered as graph nodes and are connected to their $\mathcal{K}$ most similar neighbours using cosine similarity. For each document $d_i$, we previously calculated its embedding $\mathbf{x}_i$ using Eq.~\eqref{eq:vector}, which now corresponds to node $v_i$ in the resulting graph. The KNN graph $\mathcal{G} = (\mathcal{V}, \mathcal{E})$,  consists of nodes $\mathcal{V} = \{v_1, v_2, \ldots, v_N\}$ and edges $\mathcal{E} = \{(v_i, v_j) : v_j \in \text{KNN}_\mathcal{K}(v_i)\}$, where $\text{KNN}_\mathcal{K}(v_i)$ denotes the set of $\mathcal{K}$ nearest neighbors of node $v_i$. The relationships between nodes are encoded by an adjacency matrix $\mathbf{A} \in \mathbb{R}^{N \times N}$, which stores the pairwise similarities for each KNN connection. For neighboring pairs, $A_{ij}$ is the similarity between $v_i$ and its neighbor $v_j$, $A_{ij} = 0$ for all non-neighbor pairs, and $A_{ii}$ = 1 for self loops.

\subsubsection*{Graph-based learning} Our primary goal in this stage is to obtain document embeddings that better reflect the underlying topic organization of the corpus than the initial vector representations from the first stage. To achieve this, we leverage graph-based representation learning, which means to learn representations that encode structural information about the graph. We want the resulting representations to capture a graph structure that is optimized according to our requirements. Our requirement for the graph is that documents with similar content form densely connected clusters, and each cluster corresponds to a distinct latent topic. To this end, our previously constructed KNN graph encodes the semantic relationships between documents, revealing the implicit networks. Therefore, to exploit the graph topology and generate the desired community-structured, relationally-aware embeddings, we employ Graph Convolutional Networks (GCNs). GCN is a specific type of Graph Neural Network (GNN) that uses spectral graph convolution operations and has been proven to excel at many NLP tasks \cite{yao2019graph,zhang2020graph,wu2021comprehensive}. GCN allows graph nodes to propagate information through edges and enables them to gain insight into local and global semantic patterns. The network iteratively pools information from each node's adjacent neighbors through message passing; after $n$ iterations, the embedding for each node encodes features from its $n$-hop neighborhood. Consequently, the resulting embeddings becomes influenced by their semantic neighborhood, reflecting both individual and communal semantics. While we apply GCN on the KNN graph, we additionally utilize a custom training objective\cite{zhu2021deep,zhang2020dual,wu2022hybrid} that combines hinge (pairwise ranking) and contrastive loss. This hybrid training objective drives the model to pull similar documents together and push dissimilar ones apart. With the help of this combined training objective and multiple layers of propagation, the GCN smooths out each node's document-specific noise and creates compact clusters of coherent documents. The method thus enhances discriminability between natural clusters and defines potential topic areas that facilitate smooth topic discovery. In the following paragraph, we elaborate on the GCN architecture.

\subsubsection*{Graph convolution network architecture} We employ the Cluster-GCN architecture\cite{chiang2019cluster} instead of classic GCN to handle large-scale document collections. Our Cluster-GCN approach partitions the graph into smaller sub-graphs while preserving inter-cluster edges. This partitioning strategy addresses the memory bottleneck inherent in full-batch graph processing without sacrificing model performance. The architecture consists of multiple \texttt{ClusterGCNConv} layers, with each layer performing a message-passing operation that aggregates information from neighboring nodes. Besides, it progressively reduces the dimensionality of the embeddings. The input layer accepts $D$-dimensional document embeddings from the text vectorization stage, the intermediate layers produce $d_{\text{hidden}}$-dimensional representations, and the final layer generates $d_{\text{out}}$-dimensional embeddings, yielding a dense embedding matrix. This hierarchical structure enables the model to learn compact document representations, with early layers capturing fine-grained similarities and deeper layers encoding higher-level semantic themes. Additionally, we incorporate several regularization techniques to mitigate overfitting. During each training iteration, edges are stochastically removed to prevent the model from over-relying on specific graph connections. Moreover, \texttt{GraphNorm} is applied after each convolutional layer to stabilize training by normalizing node features across the entire graph. Also, when the input and output feature dimensions are identical, residual skip connections are employed. These connections enabled direct gradient flow through the network, mitigating the vanishing gradient problem. This design facilitates training of deeper networks while preserving information from earlier layers.

\subsubsection*{Hybrid training objective} We employ a hybrid training objective for the GCN that combines two complementary loss components. This dual-objective design~\cite{joint_loss} ensures that the learned embeddings simultaneously achieve two critical goals: (1) preserving the relational structure of the documents encoded by edges in the KNN graph, through margin-based hinge loss, and (2) maintaining global semantic discriminability of the documents through contrastive loss. This section discusses how this approach helps us acquire the desired graph-based document representations.

The hinge loss operates on graph edges to reinforce the initial similarity relationships established during graph construction~\cite{hinge1,hinge2}. Through multi-layer message passing, this training objective encourages semantically similar documents (connected by an edge) to have similar embeddings, ensuring they remain closer in the learned space. Conversely, the contrastive loss enforces separation between document pairs to ensure that each document maintains a unique representation. This objective treats each node distinctively to prevent mode collapse. However, documents that share graph connections remain close despite the contrastive pressure because the hinge loss pulls them together. On the other hand, semantically dissimilar documents, lacking connecting edges, are pushed far apart by the contrastive loss. As a result, the combination of hinge and contrastive loss across multiple layers of propagation helps create distinct topic clusters in the embedding space. The loss components are described further below. 

The hinge loss component preserves local structure by ensuring that documents connected in the KNN graph (positive pairs) exhibit higher cosine similarity scores than non-neighbor documents (negative pairs), which are randomly sampled during training. The loss is activated when the similarity between a positive pair fails to exceed the similarity of a negative pair by at least a predefined margin. This margin-based formulation encourages the embedding space to separate connected documents from unconnected ones, creating clear boundaries between topical clusters. By aggregating over multiple negative samples per positive edge, the loss provides robust gradient signals that prevent the model from merely memorizing individual negative examples. 

The contrastive loss, often referred to as the NT-Xent (Normalized Temperature-scaled Cross Entropy) loss\cite{chen2020simple}, promotes global embedding distinctiveness through self-supervised discrimination. It treats each document as its own positive class, while all other documents serve as negatives. This loss function, derived from the InfoNCE (Information Noise-Contrastive Estimation) objective, encourages each document's embedding to be maximally distinguishable from all other documents in the batch. Minimizing this loss pushes the model to maximize the distance between different documents in the embedding space, ensuring that the representations capture distinctive characteristics rather than collapsing to similar values.

Prior to integration, both loss components are explicitly normalized to eliminate inherent scale variations that could otherwise bias optimization. Hyperparameter tuning across loss weighting ratios confirmed that equal weighting yields the optimal trade-off, therefore the final training objective integrates both loss terms with equal weighting:

\begin{equation} \mathcal{L} = \mathcal{L}_{\text{hinge}} + \mathcal{L}_{\text{contrast}} \end{equation}

The two losses complement one another in shaping the embedding space. Hinge loss establishes local coherence by pulling similar documents together based on graph topology, while contrastive loss establishes global discriminability by pushing dissimilar documents apart. This interplay results in an embedding space organized into well-separated semantic clusters, where within-cluster documents remain tightly grouped, and between-cluster boundaries are clearly defined. As a consequence, we get an output graph that makes the topical areas evident and separable, as shown in Fig~\ref{fig:Fig1}. Therefore, the balance between these two objectives assists us in discovering topics in the subsequent stage.

\bigskip 

The output embeddings that we get from this stage can be defined as: $\mathbf{Z} \in \mathbb{R}^{N \times d_{\text{out}}}$, which will be further processed in the next stage to identify latent topics and coherent topic words.

\subsection*{Matrix factorization}

At the third stage of GHTM, we aim to discover the latent topic structure underlying the corpus by identifying interpretable document-topic distributions. To achieve this, we perform matrix factorization on the structurally-informed document embeddings learned in the previous stage. We employ Non-negative Matrix Factorization (NMF), which decomposes the graph-enriched high-dimensional embeddings into two lower-dimensional factor matrices that reveal each document's thematic composition.

The GCN in the earlier stage produces semantically rich and structurally aware document embeddings. However, these embeddings remain in a continuous vector space without explicit topic assignments. To identify distinct topics and measure each document's affinity to those topics, we need to decompose these embeddings into a lower-dimensional latent space that captures the underlying topic structure. Traditional approaches apply NMF directly to sparse document-term matrices, where each cell represents word frequency or TF-IDF weight. However, such matrices suffer from severe lexical sparsity: documents discussing identical topics may appear dissimilar simply because they use different vocabulary (synonyms, paraphrasing, or domain-specific terminology). This approach fundamentally limits NMF's ability to discover latent topics, as the factorization operates purely on surface-level word co-occurrence patterns.

In contrast, the embeddings generated in our model’s earlier stages possess three critical properties that address these limitations while also providing additional advantages for topic identification. First, they capture semantic similarity through pre-trained GloVe representations, enabling documents with related meanings but different wording to be positioned closely in the embedding space. Second, they incorporate structural information through graph convolution, where each document's representation is refined by aggregating information from its K-nearest semantic neighbors. This message-passing mechanism amplifies consistent thematic patterns across the corpus while smoothing out noise and document-specific lexical variations. Third, the joint training objective, combining hinge loss for edge preservation and contrastive loss for discriminability organizes the embedding space into well-separated semantic clusters. This makes the latent topic boundaries more pronounced and reduces topic overlap. By factorizing these graph-enriched dense embeddings rather than raw sparse TF-IDF matrices, NMF operates on fundamentally superior document representations that reflects explicit semantic relationships and the community structure of the document embedding space. This leads to the extraction of topics characterized by more coherent word distributions and clearer thematic distinctions. The following description details the adaptation of NMF in our framework.

We begin by transforming the GCN output embeddings $\mathbf{Z} \in \mathbb{R}^{N \times d_{\text{out}}}$ into a non-negative representation, since it may contain negative values, which are incompatible with NMF requirements. Several techniques were evaluated to ensure non-negativity, including mathematical transformations (global minimum shift, min-max scaling, square, exponential, and absolute value), constrained GCN architectures, and non-negative activation functions (ReLU and SoftPlus). Among these methods, absolute value transformation yielded optimal performance in our case:

\begin{equation} 
  \mathbf{Z}_{\text{non-neg}} = |\mathbf{Z}| 
\end{equation}

This transformation preserves the magnitude of the learned relationships while maintaining strict non-negativity. Following the transformation, we decompose the transformed embedding matrix into two non-negative factor matrices with NMF:

\begin{equation}
  \mathbf{Z}_{\text{non-neg}} \approx \mathbf{W} \cdot \mathbf{H}
  \label{eq:nmf}
\end{equation}

where $\mathbf{W} \in \mathbb{R}_+^{N \times K}$ represents the document-topic association matrix and $\mathbf{H} \in \mathbb{R}_+^{K \times d_{\text{out}}}$ captures topic representations in the embedding space. Here, each row of the document-topic matrix $\mathbf{W}$, represents affinity scores over the $K$ topics for document $d_i$. Higher values indicate a stronger association with the corresponding topic, enabling us to quantify the degree to which each document belongs to each topic. This allows us to assign topics by selecting the dominant topic for each document.

Finally, to derive interpretable topic-word associations that represent the identified topics, we extract their representative documents from $\mathbf{W}$ for each topic. Then, using the original TF–IDF matrix, we sum the TF–IDF scores of each vocabulary word across those representative documents. The words with the highest cumulative scores from this summation are selected as the topic words defining the topic, as detailed in the following section.

\subsection*{Topic representation}
In this final stage, we extract interpretable topic words from the factorization results to represent the identified topics. The document-topic association matrix retrieved from NMF reveals how documents relate to topics, but does not directly indicate which words characterize each topic. To bridge this gap, we pursue the intuition that documents strongly associated with a topic should collectively exhibit vocabulary patterns which define that topic's thematic content.

For each topic $k$, we begin by identifying the documents that best represent that topic. We do that by examining the document-topic association matrix $\mathbf{W}$ from Eq.~\eqref{eq:nmf}, where each entry $W_{ik}$ indicates how strongly document $i$ is associated with topic $k$, through affinity scores. We select the top $R$ documents exhibiting the highest association scores:
\begin{equation}
    \mathcal{R}_k = \{i : W_{ik} \text{ is among top-}R \text{ values in column } k\}
\end{equation}

where $\mathcal{R}_k$ is the set of document indices that are most representative of topic $k$. These documents serve as prototypical examples of the topic, capturing its core thematic essence through their content.

Once we have identified the set of representative documents, we now determine the words that best represent the topic from these documents. The key idea is to extract the words that frequently appear across these documents which strongly belong to the topic. These words are most likely to define the topic's thematic content. To pursue this intuition, we return to the original TF-IDF matrix $\mathbf{S} \in \mathbb{R}^{N \times V}$ from Eq.~\eqref{eq:multiply}, where N is the number of documents and V is the vocabulary size of the corpus. For each word in the vocabulary, we aggregate its TF-IDF score by summing over the set $\mathcal{R}_k$ of $R$ representative documents for topic $k$.

\begin{equation}
    \boldsymbol{\omega}_k(j) = \sum_{i \in \mathcal{R}_k} S_{ij}, \quad \forall j \in \{1, 2, \ldots, V\}    %\quad \text{for } j = 1, 2, \ldots, V
\end{equation}

where $\boldsymbol{\omega}_k \in \mathbb{R}^V$ represents the aggregated term weight vector for topic $k$, and $\boldsymbol{\omega}_k(j)$ denotes the cumulative TF-IDF weight of term $j$ across all representative documents in $\mathcal{R}_k$. This operation effectively identifies the frequent and distinctive terms, from which we select the final top-$T$ topic keywords that has the highest cumulative weights for the topic:

\begin{equation}
\mathcal{T}_k = \{w_j : \boldsymbol{\omega}_k(j) \text{ is among top-}T \text{ values in } \boldsymbol{\omega}_k\}
\end{equation}

where $\mathcal{T}_k$ is the set of vocabulary terms that represent topic $k$. This distills each topic into a ranked list of the most salient terms that capture its semantic essence.

\bigskip
\bigskip
The complete GHTM framework thus combines the statistical rigor of TF-IDF, the semantic richness of pre-trained embeddings, the relational awareness of graph neural networks, and the interpretability of classical matrix factorization approaches into a unified topic modeling solution.

\section*{NCTBText dataset}
This section presents \textit{NCTBText}, a newly curated dataset comprising 8,650 text documents from textbook sources, to serve as a benchmark for Bengali topic modeling. Most of the publicly available Bengali datasets are newspaper-oriented and others are sentiment-related datasets that have been scraped from social media posts and comments. Since we suffer highly from data scarcity in Bengali, despite it being a widely spoken language, the existing datasets are extremely useful. However, texts extracted from these sources are highly biased in terms of current trends. News articles, social media posts, etc., mostly center around politics, crime, sports, and entertainment. During the development of GHTM, we noticed the absence of a standard benchmark dataset for assessing the performance of topic models in Bengali, as well as a lack of diversity in vocabulary, fresh topics, and themes in the existing datasets.

To address this gap and provide a more diverse corpus, \textit{NCTBText} is designed to serve as a reliable benchmark for future research. It comprises labeled texts from textbooks available on the Bangladesh Government’s NCTB website. The dataset encompasses diverse subjects such as Religion, Bengali Literature, Science, Agriculture, Information and Communication Technology (ICT), Business, Bangladesh and Global Studies (BGS), and Home Science, giving us a brand-new set of vocabulary for Bengali NLP. We used \textit{NCTBText} in this study for topic modeling, and we made it publicly available in \href{https://huggingface.co/datasets/farhana1996/NCTBText}{\textit{Hugging Face}} with labels so that it can be employed in any other Bengali NLP tasks. Table~\ref{tab:sample} illustrates examples of texts from \textit{NCTBText} along with their corresponding labels.

In this section, we first detail our data curation and cleaning process, followed by a description of dataset characteristics, including its class distribution and class separability in the embedding space. We also validate the dataset using Zipf’s law to demonstrate that \textit{NCTBText} exhibits the properties of natural language, and we apply several lexical diversity metrics to establish its lexical richness. Finally, we compare \textit{NCTBText}'s vocabulary with that of news-based Bengali datasets to demonstrate how our dataset broadens the lexical range by introducing diverse, domain-specific terms from textbooks.

\begin{table}[!ht]
  % \begin{adjustwidth}{-2in}{0in}
  \centering
  \small
  \caption{\textbf{Sample texts from \textit{NCTBText} with labels.}}
  \renewcommand{\arraystretch}{1.8}
  \begin{tabular}{|p{0.7\linewidth}|c|}
  \hline
  \textbf{Text} & \textbf{Label} \\
  \hline
  \bn{রাষ্ট্র, নাগরিকতা ও আইন ক্ষমতার উর্ধ্বে কোনো কর্তৃপক্ষ নেই। বাহ্যিক সার্বভৌমত্বের অর্থ হলো রাষ্ট্র আন্তর্জাতিক ক্ষেত্রে বহিশ্তির নিয়ন্ত্রণ ও হস্তক্ষেপ থেকে মুক্ত থাকবে। কেবল জনসমফ্টি, নির্দিষ্ট ভূখণ্ড ও সরকার থাকা সত্ত্বেও একটি দেশের সার্বভৌমত্ব ক্ষমতা না থাকলে তা রাষ্ট্র হিসেবে বিবেচিত হবে না। যতদিন রাষ্ট্রের স্থায়িত্ব বিদ্যমান থাকে ততদিন সার্বভৌমত্তরের স্থায়িত্ব থাকবে৷ সরকারের পরিবর্তন সার্বভৌমত্বের স্থায়িত্বকে নষ্ণ করে না। সুতরাং, জনসমফ্টি , নির্দিষ্ট ভূখন্ড, সরকার এবং সার্বভৌমত্ব এ চারটি উপাদান নিয়েই রাষ্ট্র গঠিত হয়। এর যে কোনো একটি উপাদান না হলে রাষ্ট্র গঠিত হতে পারে না।} & BGS \\
  \hline
  \bn{মাধ্যম ভিত্তিক নার্সারি আবার দুই ধরনের পলিব্যাগ নার্সারি এ ধরনের নার্সারিতে পলিব্যাগে বীজ বপন করে চারা উৎপাদন করা হয় । পলিব্যাগ সহজে সরানো যায় বলে চারাকে খরা, বৃষ্টি ও দুর্যোগ থেকে রক্ষা করা যায় । গাছ থেকে গাছে রোগ সংক্রমণ কম হয় ৷ এ পদ্ধতিতে নিবিড়ভাবে চারার যত্ন নেওয়া যায় । বেড নার্সারি নার্সারি তৈরির এ পদ্ধতিতে সরাসরি মাটিতে বেড তৈরি করে চারা উৎপাদন করা হয় ।} & Agriculture \\
  \hline
  \bn{ই-বুক ই-বুক বা ইলেক্ট্রনিক বুক বা ই-বই হলো মু্রিত বইয়ের ইলেক্ট্রনিক রূপ। যেহেতু, এটি ইলেকট্রনিক মাধ্যমে প্রকাশিত হয় সে কারণে এতে শব্দ, এনিমেশন ইত্যাদিও জুড়ে দেওয়া যায় । অবশ্য এখন অনেক ই-বুক কেবল ই-বুক আকারে প্রকাশিত হয়৷ এগুলোর মুদ্রিত রূপ থাকে না। ফলে অনেকেই এখন আর ই-বুককে মুদ্রিত বইয়ের ইলেক্ট্রনিক সংস্করণ বলতে নারাজ ৷} & ICT \\
  \hline
  \end{tabular}
  \begin{flushleft}
  Text samples from the \textit{NCTBText} dataset with corresponding domain labels. The text documents are lengthy because they cover one page of textbook data. For improved readability, texts have been shortened here. Here, BGS = Bangladesh and Global Studies and ICT = Information and Communication Technology.
  \end{flushleft}
  \label{tab:sample}
  % \end{adjustwidth}
  \end{table}

\subsection*{Data curation}
We collected the textbooks from the official NCTB platform, which distributes free e-books of the national curriculum for primary, secondary, and higher secondary education in Bangladesh. As these materials are publicly available, their collection and use for research purposes comply with the terms of their public distribution. We gathered the PDFs of these e-books for selected subjects from grades 6 to 12, published for the year 2025. We transformed the book pages into images utilizing the Python module \texttt{pdf2image} and performed OCR with \texttt{pytesseract} to extract Bengali text from the images. Compared to the other tools available for Bengali text OCR, only \texttt{pytesseract} performed well. We stored one page of text as a single text document in our corpus, where the label is set according to the book’s subject.

\subsection*{Data cleaning} 
We did rigorous and careful data cleaning of the \textit{NCTBText} by removing unnecessary OCR artifacts, English texts, numerics, redundant spaces, line breaks, etc. As the curated data is from textbooks, there were no emojis, URLs, or emails to remove, but there were a lot of single vowels (e.g.,\bn{ ী , ি , ু , া }), letters (e.g., \bn{ক, খ, গ, ঘ}), and unnecessary punctuations that we removed as a part of the cleaning process. The cleaned text was then normalized to \href{https://unicode.org/reports/tr15/}{NFKC (Normalization Form KC)} encoding. The classes, or in this case subjects, that had less than five hundred documents, such as \textit{Work and Career}, \textit{Arts and Crafts}, \textit{Physical Education}, etc., were excluded from the dataset to ensure sufficient representation of each class.

\subsection*{Data characteristics} 
This subsection details the dataset's class distribution and the methodology used to derive these classes from book subjects, which is essential for understanding the dataset’s characteristics and balance. Additionally, since these categories were custom-defined, we assess their separability in the embedding space to validate the dataset's viability for classification.

\subsubsection*{Class distribution}  
\begin{figure}[!h]
    \centering
    \includegraphics[width=0.7\textwidth]{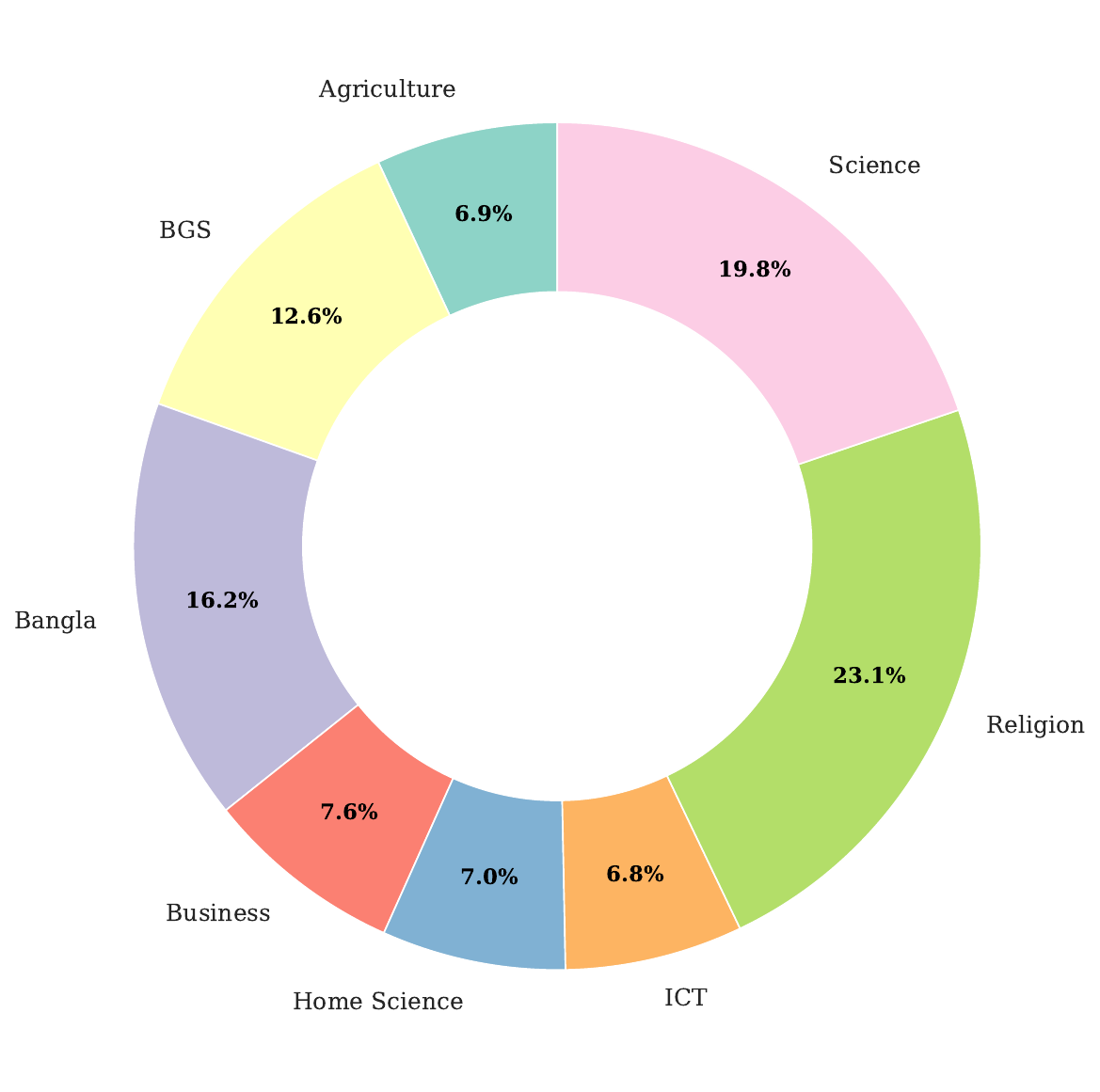}
    \caption{\textbf{Class Distribution of \textit{NCTBText}.}
  The 8,650 texts in the dataset are distributed across eight subject categories as follows, in descending order: Religion (1,999), Science (1,712), Bengali (1,399), Bangladesh and Global Studies - BGS (1,090), Business (656), Home Science (607), Agriculture (600), and Information and Communication Technology - ICT (587).}
  
  %The dataset consists of 8650 texts that are distributed across eight subject categories. Religion comprises 1999 texts, Science comprises 1712, Bengali comprises 1399, Bangladesh and Global Studies (BGS) comprises 1090, Business comprises 656, Home Science comprises 607, Agriculture comprises 600, and Information and Communication Technology (ICT) comprises 587.}
  
    \label{fig:Fig2}
\end{figure}

The distribution chart in Fig~\ref{fig:Fig2} illustrates the class distribution of the \textit{NCTBText} dataset. The largest portion belongs to Religion, which comprises 1,999 text documents. This category merges Muslim, Hindu, Christian, and Buddhist texts, as each of these textbooks individually generated fewer than five hundred documents. We therefore consolidated them under the unified theme of religion.

The second largest category is Science. For grades 6–8, NCTB provides a general science textbook encompassing all major branches of science. However, for grades 9–10, a general science textbook is available only for commerce students, while science students use separate textbooks for \textit{Physics}, \textit{Chemistry}, and \textit{Biology}. To maintain consistency, we combined these specialized branches with the general science texts into a single Science category.

Similarly, for the BGS (Bangladesh and Global Studies) category, we integrated the general BGS texts for grades 6–8 with the specialized \textit{Geography}, \textit{History}, and \textit{Civics} textbooks for grades 9–10. The Bengali category compiles literature books from grades 6–12, encompassing essays, poetry, novels, plays, and short stories.

The Business category unifies the \textit{Accounting}, \textit{Finance and Banking}, and \textit{Business Entrepreneurship} textbooks from grades 9–10. In contrast, \textit{Home Science} and \textit{Agriculture Studies} textbooks (grades 6–10) and \textit{ICT} books (grades 6–12) are provided as general textbooks without subject-specific branches; thus, we only aggregated texts across grade levels for these subjects.

\subsubsection*{Class separability} 
\begin{figure}[!h]
  \centering
  \includegraphics[width=0.8\textwidth]{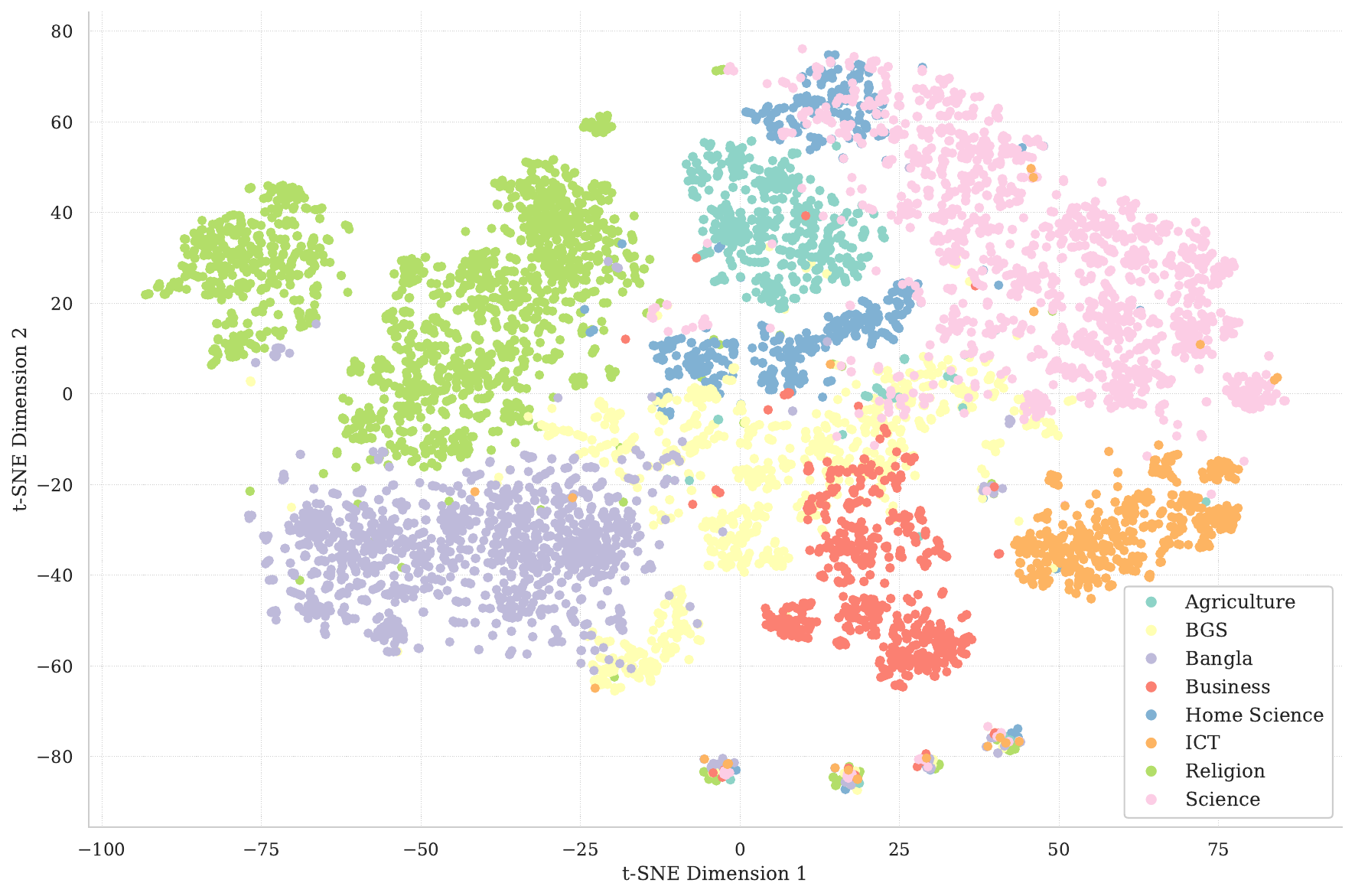}
  \caption{\textbf{Two-dimensional t-SNE projection of \textit{NCTBText}.}
  The document embeddings of the \textit{NCTBText} dataset are represented in two dimensions using t-SNE. Each point depicts a document, colored in accordance with its true class label. Distinct clusters show that the categories are well-separable in the embedding space.}
  \label{fig:Fig3}
  \end{figure}

In order to assess the separability of the categories and demonstrate that these classes are distinctive enough for classification, we visualize the documents of \textit{NCTBText}. We generate document embeddings with Bengali SBERT model\cite{uddin2024bangla} and plot them using t-SNE\cite{vandermaaten2008visualizing}. The resulting plot is depicted in Fig~\ref{fig:Fig3}. As shown in the figure, the clusters are distinctly defined and separable, except for Home Science, which is slightly conflated with Science. This phenomenon occurs because several Home Science chapters concerning food and nutrition closely resemble the biology chapters in Science. Moreover, we can observe that the t-SNE plot reflects the class distribution. The Science and Religion group predominates the map without overshadowing the minor clusters. The four small, jumbled data clusters in the lower section of the map comprise common texts prevalent in all textbooks, featuring preface terms such as \bn{``পাঠ্যপুস্তক", ``পরিমার্জিত", ``শিক্ষাবর্ষ"} etc. We did not remove these terms because real-world data is inherently messy, and removing them would compromise the dataset's authenticity.

\subsection*{Data validation}
Beyond structural analysis, we validate \textit{NCTBText} in this subsection against quantitative measures. We demonstrate that the dataset adheres to natural language patterns, as evidenced by its compliance with Zipf's Law, and exhibits substantial lexical richness, as measured by multiple diversity metrics.

\subsubsection*{Zipf’s law} 
\begin{figure}[!h]
  \centering
  \includegraphics[width=\linewidth]{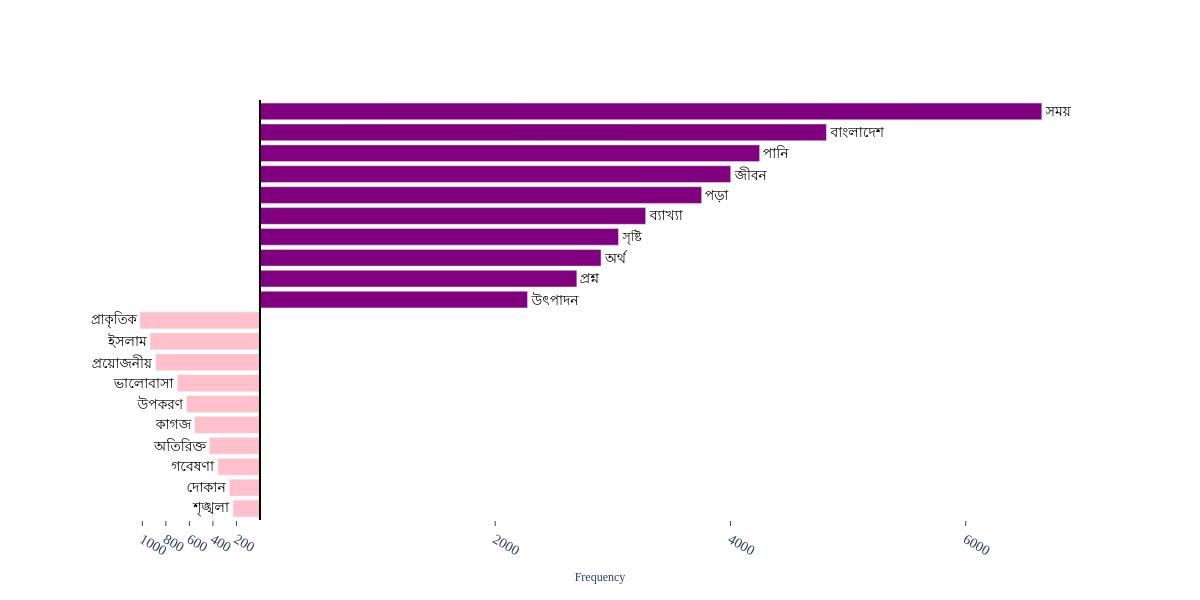}
  \caption{\textbf{Common and Rare Words in \textit{NCTBText}.} 
  Frequency distribution of the common and rare words in the \textit{NCTBText} dataset. Common words that are frequently found on the dataset are represented in purple on the right, while the words with low frequencies are represented in pink on the left.}
  \label{fig:Fig4}
\end{figure}

To assess the language properties of our dataset, we validate it against Zipf's Law~\cite{Zipf1935}. Zipf's Law states that in natural language corpora, word frequency is inversely proportional to rank, following a power-law distribution. Mathematically, this relationship is expressed as $f(r) = \frac{C}{r^s}$, where $f(r)$ denotes the frequency of the word at rank $r$, $C$ is a normalization constant, and $s$ is the Zipf exponent. This distribution reflects a fundamental property of natural language: a small set of common words appears with high frequency, while the vast majority of words occur rarely. Fig~\ref{fig:Fig4} presents an overview of this phenomenon, highlighting the frequencies of some common and rare words in the NCTBText vocabulary.

\begin{figure}[!h]
  \centering
  \includegraphics[width=0.7\textwidth]{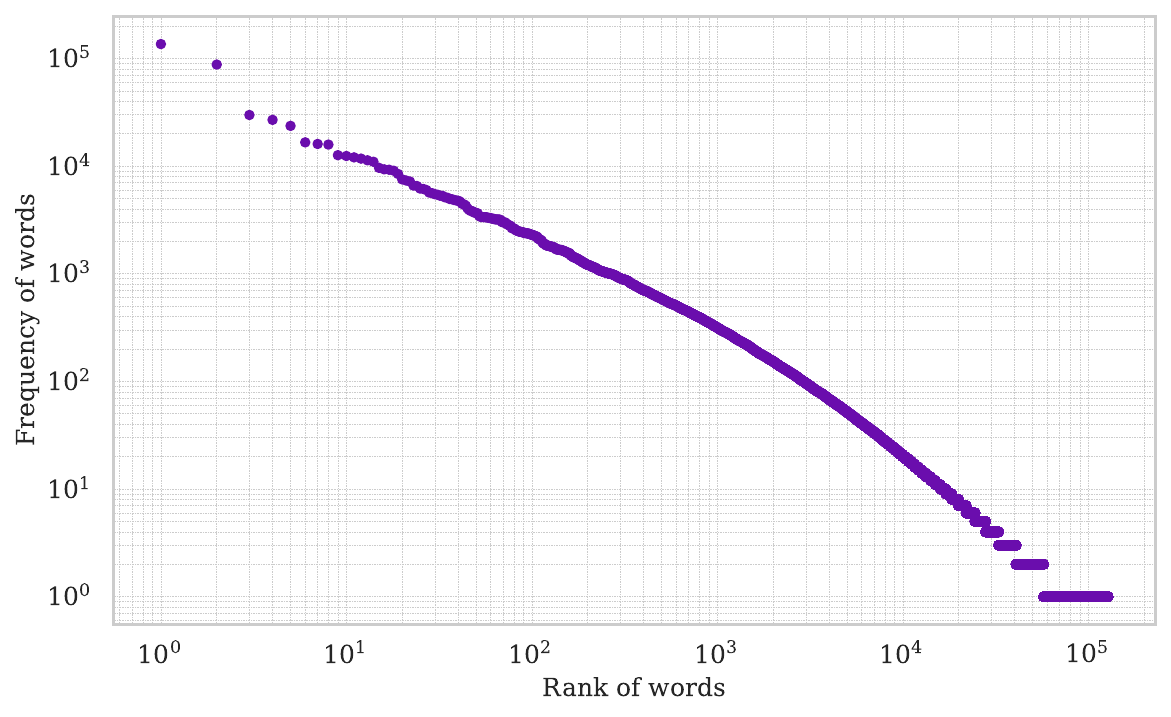}
  \caption{\textbf{Zipf’s Law Validation of NCTBText.}
  Log-log graph of word frequency against rank. In accordance with Zipf’s law, this distribution indicates that in the \textit{NCTBText} dataset, a small number of words occur very frequently while the majority of words appear rarely.}
  \label{fig:Fig5}
  \end{figure}

To validate \textit{NCTBText} in accordance with this principle, we plotted the logarithm of word frequencies against the logarithm of word ranks. As shown in Fig~\ref{fig:Fig5}, the resulting distribution exhibits a nearly linear relationship, confirming that \textit{NCTBText} adheres to Zipf's Law. This compliance demonstrates that our dataset preserves the natural statistical properties of language, validating its authenticity for NLP applications.

\subsubsection*{Lexical diversity} 

To verify the linguistic richness of our dataset, we performed a thorough analysis utilizing multiple lexical diversity metrics. Lexical diversity~\cite{malvern2004lexical} quantifies the range and variety of vocabulary in a text or corpus, serving as a key indicator of linguistic quality and dataset suitability for natural language processing applications. We calculated lexical diversity metrics for our dataset, including TTR (Type-Token Ratio), MTLD (Measure of Textual Lexical Diversity), and MATTR (Moving-Average Type-Token Ratio)\cite{mccarthy2010, covington2010, fergadiotis2015}. This multi-metric methodology facilitates an extensive evaluation of vocabulary richness of the \textit{NCTBText} dataset. The metrics employed in this study are briefly introduced below.

\textbf{Type-Token Ratio (TTR)}~\cite{carroll1964language, malvern2004lexical} is the foundational metric for assessing lexical diversity, that compares the number of unique words (types) to the total number of words (tokens) in a text. Although TTR offers an intuitive assessment of vocabulary richness, it exhibits a fundamental limitation that is, strong sensitivity to text length. As the length of a document increases, TTR typically decreases because common vocabulary is repeated more often. Another metric called \textbf{Measure of Textual Lexical Diversity (MTLD)}\cite{mccarthy2005assessment},\cite{mccarthy2010} addresses this issue. MTLD calculates lexical diversity by sequentially processing tokens and measuring segment lengths. This approach produces length-independent diversity estimates by normalizing for the natural decline in TTR as text length increases. \textbf{Moving-Average Type-Token Ratio (MATTR)} \cite{covington2010}, represents another robust, length-independent diversity metric. MATTR operates by computing TTR over consecutive, overlapping windows of fixed size (typically 100 tokens). The results of this evaluation are presented in Table~\ref{tab:lexical_diversity}, which shows the mean, standard deviation (SD), minimum (Min), and maximum (Max) values of each metric across all documents in the dataset. These statistics reveal important characteristics of the linguistic composition of our dataset.

\begin{table}[!ht]
  \centering
  \caption{\textbf{Lexical Diversity Results for \textit{NCTBText}.}}
  \renewcommand{\arraystretch}{1.8}
  \begin{tabular}{|l|c|c|c|c|}
  \hline
  \textbf{Metric} & \textbf{Mean} & \textbf{SD} & \textbf{Min} & \textbf{Max} \\
  \hline
  TTR   & 0.6296  & 0.0970  & 0.1579   & 1.0000 \\
  \hline
  MTLD  & 94.4286 & 52.4080 & 3.2423   & 479.1667 \\
  \hline
  MATTR & 0.7848  & 0.1366  & 0.0000   & 0.9800 \\
  \hline
  \end{tabular}
  \begin{flushleft}
  %Lexical diversity analysis of \textit{NCTBText} dataset showing mean, standard deviation (SD), minimum, and maximum values for Type-Token Ratio (TTR), Measure of Textual Lexical Diversity (MTLD), and Moving Average Type-Token Ratio (MATTR).
  \end{flushleft}
  \label{tab:lexical_diversity}
\end{table}

The TTR measures in Table~\ref{tab:lexical_diversity} reflect considerable lexical richness, with an average of 62.96\%, indicating that nearly two-thirds of tokens represent unique word types. This is particularly noteworthy given that \textit{NCTBText} consists of lengthy documents. This shows that our dataset encompasses diverse vocabulary rather than redundant content. On the other hand, the high mean MTLD score of 94.43 for \textit{NCTBText} indicates strong lexical diversity across the dataset. However, the large standard deviation and the wide range of MTLD values (3.24–479.17) reflect the topical diversity within the corpus, encompassing documents from varying domains and complexity levels. Similarly, MATTR measurements of \textit{NCTBText} from Table~\ref{tab:lexical_diversity} also provide strong evidence for lexical richness across the corpus.

These results collectively demonstrate that \textit{NCTBText} exhibits substantial lexical richness. This diversity in vocabulary validates the dataset's suitability for downstream NLP applications that require rich vocabulary coverage. \textit{NCTBText}'s wide range of linguistic styles and domains makes it a valuable resource for Bengali NLP research.

\subsection*{Beyond newspaper dataset: expanding the lexical range for Bengali NLP} 
\begin{figure}[!h]
  \includegraphics[width=\linewidth]{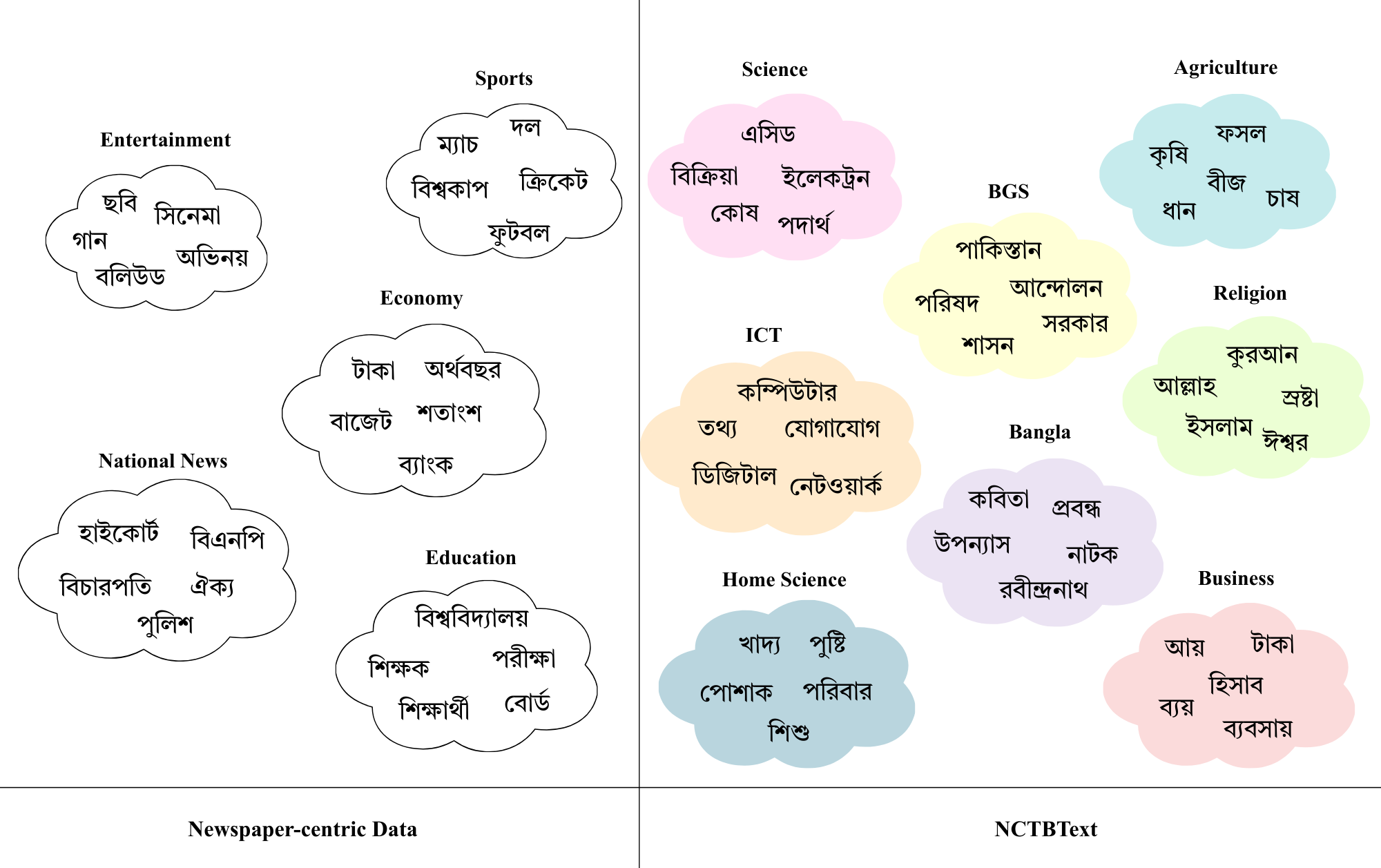}
  \caption{\textbf{News vs. \textit{NCTBText} Vocabulary.}
  Left : Top words formed in word clouds from five common categories found in Bengali newspaper datasets. Right: Prevalent words identified in \textit{NCTBText} classes formed in word clouds.}
\label{fig:Fig6}
\end{figure}

To conduct a thorough assessment of topic models in Bengali, we needed datasets that varied in size, nature, document-length and vocabulary. Unfortunately, we could only find datasets derived from newspaper websites and social media, characterized by vocabulary centered on trending topics. These datasets undoubtedly alleviate the data scarcity issue in Bengali, yet the vocabulary remains limited to journalistic and editorial conventions, lacking terminological depth. We reviewed several datasets in our quest of benchmark datasets for topic model evaluation, including \textit{Shironaam}\cite{akash2023shironaam}, \textit{Potrika}\cite{ahmad2022potrika}, \textit{BNAD}\cite{saad2024bnad}, and \textit{BanFakeNews}\cite{hossain2020banfakenews}. To compare their lexical range with \textit{NCTBText}, we selected five prevalent categories from these datasets: National News, Education, Sports, Entertainment, and Economy. Fig~\ref{fig:Fig6} illustrates the disparity in vocabulary by highlighting the most prevalent words identified in newspaper datasets and in \textit{NCTBText}. The visualization reveals that \textit{NCTBText} being curated from textbooks, added more diversity and richness in Bengali NLP vocabulary by introducing subject-specific jargons and transliterations such as \bn{``এনজাইম"} (Enzyme), \bn{``ইলেকট্রন"} (Electron), \bn{``নেটওয়ার্ক"} (Network) etc. This inclusion of academic terminologies represents a significant expansion of Bengali computational linguistics resources, bridging the gap between everyday language use and specialized knowledge domains.

\section*{Experimental setup}
To ensure reproducibility and provide a transparent basis for comparing topic models, this section details our experimental framework, covering the datasets, data preparation, baseline models, evaluation metrics, model and environment configurations, and hyperparameter settings.

\subsection*{Datasets}
This research utilizes three Bengali datasets of varying sizes and one English benchmark dataset to evaluate the proposed GHTM approach against existing topic modeling techniques. The datasets are briefly introduced below:

\textbf{Jamuna News} is curated from the Jamuna TV website, a national news broadcasting channel in Bangladesh. It is a balanced collection of short documents in Bengali, obtained from \href{https://www.kaggle.com/datasets/durjoychandrapaul/over-11500-bangla-news-for-nlp}{\textit{Kaggle}}.

\textbf{BanFakeNews}\cite{hossain2020banfakenews}, is compiled from Bengali newspaper articles. The dataset was released to combat the spread of fake news in Bengali. In this study, we use its \textit{``Authentic-48K"} subset which contains the authentic news only.

\textbf{NCTBText} is the novel Bengali dataset that we introduced in the previous section.

\textbf{20NewsGroup}\cite{twenty_newsgroups_113} is a widely used benchmark collection for English, which contains posts from the Usenet newsgroups of the mid-1990s. %It is a very popular dataset that encompasses diverse themes such as politics, religion, sports, science, and technology.

\bigskip

The attributes of the datasets are presented as a summary in Table~\ref{tab:dataset_summary}.

\begin{table}[!ht]
  \centering
  \caption{\textbf{Dataset summary.}}
  \renewcommand{\arraystretch}{1.8}
  \begin{tabular}{|l|c|c|c|c|c|}
  \hline
  %\textbf{Dataset} & \textbf{Documents} & \textbf{Vocabulary} & \textbf{Classes} & \textbf{Size} & \textbf{Avg. Words} \\
  \textbf{Dataset} & \textbf{\thead{Total\\Documents}} & \textbf{\thead{Vocabulary\\Size}} & \textbf{\thead{No. of\\Classes}} & \textbf{\thead{Storage\\Size}} & \textbf{\thead{Avg.\\Document\\Length}} \\
  \hline
  Jamuna News & 11,904 & 34,101 & 4 & 19.2 MB & 89.00 \\
  \hline
  NCTBText & 8,650 & 84,269 & 8 & 37.6 MB & 271.73 \\
  \hline
  BanFakeNews & 48,678 & 130,227 & 12 & 244.4 MB & 304.59 \\
  \hline
  20NewsGroup & 18,846 & 110,993 & 20 & 46.7 MB & 93.36 \\
  \hline
  \end{tabular}
  \begin{flushleft}
  Dataset characteristics including total document count, vocabulary size, number of ground truth classes, storage size (before cleaning), and average word count per document (document-length) .
  \end{flushleft}
  \label{tab:dataset_summary}
  \end{table}

\subsection*{Data preparation}
We prepared the datasets in both tokenized sequences and raw sentence forms to accommodate the diverse requirements of models used in the comparative analysis. Different models had different input requirements, which could be generalised into these two formats. While some models accept raw sentences as they work with sentence-level embeddings, other conventional and neural models accept a tokenized list of words for each document. Certain models, on the other hand, leverage both formats. For GHTM specifically, we used the tokenized format since our text representation pipeline begins with the synthesis of TF-IDF and word-level embeddings, which operate on tokenized word sequences rather than sentence embeddings. For models requiring tokenized input, we applied rigorous preprocessing steps, including tokenization, stop-word removal, and lemmatization. And for models requiring raw sentences, we applied minimal preprocessing, removing only unnecessary punctuation and numerics to preserve sentence structure. 

\subsection*{Baselines}
In the related work section, we discussed the evolution of topic models, outlining the fundamental architectural categories. None of the significant models from these categories, except LDA, have been applied on Bengali for performance evaluation. Therefore, to conduct a comprehensive analysis of existing topic modeling methods on Bengali datasets, we employed a wide range of baseline models. Our baseline selection encompasses classic models (LDA, LSA, NMF), neural topic models (ProdLDA, NeuralLDA, ETM, CombinedTM, ZeroShotTM), cluster-based approaches (Top2Vec, BERTopic, CluWords), and graph-based architectures (\href{https://github.com/SmilesDZgk/GNTM/tree/master}{GNTM}, \href{https://github.com/zhehengluoK/GCTM}{GCTM}, \href{https://pypi.org/project/graph2topictm/}{Graph2Topic}), ensuring broad coverage of the topic modeling landscape.

However, we encountered a few challenges while implementing the baselines. First, we were unable to reproduce results for \href{https://github.com/AdhyaSuman/GINopic/tree/master}{GINopic} and \href{https://github.com/valdersoul/GraphBTM/tree/master}{GraphBTM} on Bengali datasets, as their respective implementations did not include explicit guidance for data preprocessing requirements for any language other than English. Second, Bengali-specific topic models such as BERT-LDA, LCD and the clustering-based approach\cite{rifat2025clustering} could not be included as baselines despite their relevance to our research due to unavailable public repositories or implementation. Third, the representative of LLM-based models, TopicGPT, generates topic labels and descriptions rather than the lists of topic words that traditional evaluation metrics require. Hence, TopicGPT's output format renders it incompatible with our evaluation framework, ruling out its inclusion as a baseline.

On the other hand, for the performance comparison on the English benchmark dataset, we solely used graph-based models as baselines, including GraphBTM, GNTM, Graph2Topic, GCTM, and GINopic.

\subsection*{Configuration and setup}
All the baseline models employed in the study were set to their default configurations and hyperparameters for the comprehensive evaluation. However, we had to modify \texttt{sklearn}'s \texttt{CountVectorizer} tokenization pattern for the models that uses this module for vectorization, to accurately handle Bengali text tokenization. 

The baseline models for Bengali used a variety of embedding models throughout the experiment, which are summarized in Table~\ref{tab:embeddings_model_summary}. For the English performance evaluation, we used each baseline's default embedding model, while for GHTM, we used \href{https://nlp.stanford.edu/projects/glove/}{\texttt{glove.840B.300d.txt}}.

All experiments were conducted using a combination of local and cloud-based GPU environments. The local experiments were executed on a laptop equipped with an NVIDIA GeForce RTX 4060 GPU (8 GB). For more computationally demanding models and larger datasets, cloud environments were employed. This included \href{https://www.kaggle.com/}{\textit{Kaggle}’s} dual NVIDIA Tesla T4 GPUs (2 × 16 GB) and \href{https://lightning.ai/}{\textit{Lightning AI}’s} NVIDIA L4 GPU (24 GB) instances.

\begin{table}[!ht]
  \centering
  \caption{\textbf{Embedding models summary.}}
  \renewcommand{\arraystretch}{1.5}
  \begin{tabular}{|l|l|c|}
  \hline
  \textbf{Embedding} & \textbf{Model Name} & \textbf{Dimension} \\
  \hline
  Word2Vec & \texttt{bnwiki\_word2vec} & 100 \\
  \hline
  Doc2Vec & \texttt{bangla\_news\_article\_doc2vec} & 100 \\
  \hline
  GloVe & \texttt{bn\_glove.39M.300d} & 300 \\
  \hline
  FastText & \texttt{fasttext\_cc.bn.300} & 300 \\
  \hline
  SBERT\cite{uddin2024bangla} & \href{https://huggingface.co/shihab17/bangla-sentence-transformer}{\texttt{bangla-sentence-transformer}} & 768 \\
  \hline
  \end{tabular}
  \label{tab:embeddings_model_summary}
\end{table}

\subsection*{Hyperparameter configuration}
In this section, we discuss the hyperparameters of GHTM and their configuration, which were tuned to balance model performance with computational constraints. Table~\ref{tab:hyperparameters} presents the hyperparameter configurations used for GHTM across the datasets.

\begin{table}[!ht]
  % \begin{adjustwidth}{-2in}{0in}
  \centering
  \small
  \caption{\textbf{GHTM hyperparameter values.}}
  \renewcommand{\arraystretch}{1.5}
  \begin{tabular}{|l|c|c|c|c|}  
  \hline
  \textbf{Hyperparameter} & \textbf{Jamuna News} & \textbf{NCTBText} & \textbf{BanFakeNews} & \textbf{20NewsGroup}\\
  \hline
  No. of Topics, $K$ & 4 & 8 & 12 & $K \in {10, 20, 50}$ \\
  \hline
  Hidden Layers & 5 & 4 & 1 & 1\\
  \hline
  GCN Clusters, $P$ & 2 & 4 & 8 & 8\\
  \hline
  Hidden Layer Dimension, $d_{\text{hidden}}$ & 32 & 32 & 128 & 256\\
  \hline
  Output Layer Dimension, $d_{\text{out}}$ & \multicolumn{4}{c|}{64} \\ 
  \hline
  Epochs & \multicolumn{4}{c|}{100} \\  
  \hline
  KNN Neighbors, $k$ & \multicolumn{4}{c|}{15} \\ 
  \hline
  Learning Rate & \multicolumn{4}{c|}{0.005} \\ 
  \hline
  Dropout & \multicolumn{4}{c|}{0.4} \\ 
  \hline
  Edge Dropout & \multicolumn{4}{c|}{0.2} \\  
  \hline
  Representative Documents, $R$  & \multicolumn{4}{c|}{20} \\  
  \hline
  Top Keywords per Topic, $T$ & \multicolumn{4}{c|}{10} \\  
  \hline
  \end{tabular}
  \label{tab:hyperparameters}
% \end{adjustwidth}
\end{table}

For Bengali datasets, we set the number of topics $K$ to the ground truth values to ensure fair and straightforward comparison across baselines in terms of topic coherence and diversity. For English datasets, we report averaged topic coherence scores across $K \in \{10, 20, 50\}$.

Hyperparameter tuning revealed systematic patterns reflecting the interplay between dataset characteristics and model architecture. As dataset size and average document length increased, performance improved with shallower GCN architectures (fewer hidden layers) but required larger hidden dimensions. This inverse relationship suggests that larger datasets benefit from greater representational capacity per layer rather than architectural depth, while deeper networks are more effective for shorter documents where smoothing is necessary for a stable representation.

On the other hand, the number of clusters $P$ in Cluster-GCN critically affects the trade-off between memory consumption and computational efficiency. Smaller $P$ values (larger clusters) provide richer neighborhood context per batch but demand substantially more GPU memory, making them feasible only with sufficient hardware resources. Conversely, larger $P$ values (smaller clusters) reduce memory requirements, enabling training on resource-constrained devices, though at the cost of reduced model performance. Therefore, based on these observations we selected all our hyperparameters that maximized model performance, while balancing GPU memory constraints and dataset attributes.

\subsection*{Evaluation metrics}
We evaluate the performance of the models in terms of both topic coherence and topic diversity. For coherence evaluation, we employ NPMI and $\mathbf{C_V}$, whereas topic diversity is measured through TD and IRBO. A brief overview of these metrics is given below.

\textbf{Normalized Pointwise Mutual Information (NPMI)}\cite{newman-etal-2010-automatic} is a statistical measure of word association that measures topic coherence using a sliding window to count word co-occurrence patterns. The measure ranges from [-1, 1] where 1 indicates a perfect relevance of words in a topic. The concept is represented as:

\begin{equation}
\mathrm{NPMI}\left(w_x,w_y\right)=\frac{\log \frac{P\left(w_x,w_y\right)}{P\left(w_x\right)P\left(w_y\right)}}{-\log P\left(w_x,w_y\right)}
\end{equation}
where $P(w_x)$ and $P(w_y)$ are probabilities of the words $w_x$ and $w_y$ and $P(w_x,w_y)$ is the joint probability of co-occurrence.

\textbf{Topic Coherence ($\mathbf{C_V}$)}\cite{roder2015exploring} is a variant of NPMI that measures the semantic relatedness of topic words. It uses the one-set segmentation to count word co-occurrences and the cosine similarity as the similarity measure. It ranges from [0, 1], where 1 means closely related words identified as topic representatives. 

\textbf{Topic Diversity (TD)}\cite{dieng2020topic} measures the uniqueness of the words across all topics and the measure ranges within [0, 1] where 0 indicates redundant and overlapping topics and 1 indicates highly diverse topics. It can be mathematically expressed as:

\begin{equation}
  \mathrm{TD}
    = \frac{\bigl|\bigcup_{k=1}^K \mathcal{T}_k\bigr|}{K \times T}
\end{equation}
where $\mathcal{T}_k$ represents the set of top $T$ words for topic $k$, and $K$ is the total number of topics.

\textbf{Inverted Rank-Biased Overlap (IRBO)}\cite{webber2010similarity}, a diversity metric that evaluates inter‐topic dissimilarity, derived from Rank-Biased Overlap (RBO). While RBO quantifies the similarity of ranked word lists across topics, IRBO inverts RBO to penalize overlapping top words. 

We used \texttt{CoherenceModel} from \textit{Gensim}\cite{rehurek2010software} to compute coherence metrics (NPMI and $\mathbf{C_V}$) and \textit{OCTIS (Optimizing and Comparing Topic models is Simple)}\cite{terragnioctis2021} to calculate diversity metrics (TD and IRBO). 

\textbf{Runtime (RT)} is also recorded for each run to observe how long models take to train and generate topics as the size of the dataset grows.

\section*{Results and analysis}
This section presents a comprehensive evaluation of topic modeling approaches on Bengali datasets, ranging from traditional probabilistic and algebraic models to modern neural, cluster-based, and graph-based architectures. In addition, we report the results of contemporary graph-based models on the English benchmark dataset \textit{20NewsGroup}. Beyond the quantitative comparisons, we also qualitatively assess topics through topic quality analysis. Furthermore, we conduct an ablation study and a statistical significance test to validate our findings and provide deeper insights. 

\subsection*{Comprehensive evaluation of topic models on Bengali}
We selected a wide range of baselines, datasets, and evaluation metrics to conduct a thorough evaluation of the topic models on Bengali datasets. Table~\ref{tab:full_results} provides a holistic view of the evaluation results. In the following paragraphs, we discuss the performances by model category.

Amidst the \textbf{traditional models}, NMF with TF-IDF vectorization emerged as the most effective approach, achieving balanced performance ($\mathbf{C_V}$: 0.57–0.72, TD: 0.94–0.98) with notably fast runtime (0.6–15.86 seconds). In contrast, LSA consistently underperformed across all datasets, with particularly low coherence ($\mathbf{C_V}$: 0.34–0.51) and diversity scores (TD: 0.39–0.68). Although LDA and NMF achieved reasonable coherence scores, they exhibited critical limitations on \textit{BanFakeNews}. Both models identified only 10 topics despite the number of topics, $K$, being set to 12, indicating inadequate topic discovery capacity for large datasets.

Among the \textbf{Neural Topic Models (NTMs)}, CombinedTM achieved the highest topic coherence ($\mathbf{C_V}$: 0.64–0.71) and perfect diversity (TD: 1.00), demonstrating overall competitive performance across all models. The model leverages both BoW and SBERT embeddings, which signifies the advantage of hybrid text representations in topic modeling. In contrast, ETM demonstrated subpar performance compared to its counterparts, despite employing FastText for text vectorization, which is typically effective for morphologically intricate languages like Bengali. Moreover, the high runtimes (13.63–2354.88 seconds) observed across NTMs without corresponding performance gains limit their practical applicability.

\begin{table}[H]
  \begin{adjustwidth}{-0.5in}{0in}
  \centering
  \scriptsize
  \caption{\textbf{Comprehensive model comparison across Bengali datasets.}}
  \renewcommand{\arraystretch}{1.5}
  \begin{tabular}{|>{\raggedright}p{3.5cm}|*{15}{c|}}
  \hline
  \multirow{2}{*}{\textbf{Model}} 
  & \multicolumn{5}{c|}{\textbf{Jamuna News}} 
  & \multicolumn{5}{c|}{\textbf{NCTBText}} 
  & \multicolumn{5}{c|}{\textbf{BanFakeNews}} \\
  \cline{2-16}
  & \specialcell{$\mathbf{C_V}$} 
  & \specialcell{NP\\MI} 
  & \specialcell{TD} 
  & \specialcell{IR\\BO} 
  & \specialcell{RT} 
  & \specialcell{$\mathbf{C_V}$} 
  & \specialcell{NP\\MI} 
  & \specialcell{TD} 
  & \specialcell{IR\\BO} 
  & \specialcell{RT} 
  & \specialcell{$\mathbf{C_V}$} 
  & \specialcell{NP\\MI} 
  & \specialcell{TD} 
  & \specialcell{IR\\BO} 
  & \specialcell{RT} \\
  \hline
  
  \multicolumn{16}{|l|}{\textbf{Traditional Models}} \\
  \hline
  LDA (BOW) & 0.62 & 0.08 & 0.89 & 0.95 & 2.44 & 0.50 & 0.03 & 0.83 & 0.95 & 2.43 & 0.64 & 0.12 & 0.86 & 0.97 & 32.38 \\
  \hline
  LDA (TF-IDF) & 0.57 & 0.03 & 0.91 & 0.95 & 2.15 & 0.51 & -0.04 & 0.96 & 0.99 & 2.04 & 0.65 & 0.09 & 0.97 & 1.00 & 29.88 \\
  \hline
  LSA (BOW) & 0.51 & -0.04 & 0.60 & 0.63 & 0.39 & 0.34 & -0.05 & 0.47 & 0.78 & 0.72 & 0.44 & -0.01 & 0.39 & 0.79 & 11.81 \\
  \hline
  LSA (TF-IDF) & 0.49 & -0.05 & 0.68 & 0.67 & 0.73 & 0.42 & -0.07 & 0.61 & 0.78 & 1.29 & 0.45 & -0.03 & 0.42 & 0.83 & 20.78 \\
  \hline
  NMF (BOW) & 0.62 & 0.09 & 0.91 & 0.95 & 0.30 & 0.54 & 0.06 & 0.82 & 0.96 & 0.38 & 0.67 & 0.13 & 0.84 & 0.97 & 7.84 \\
  \hline
  NMF (TF-IDF) & 0.65 & 0.08 & 0.98 & 0.99 & 0.60 & 0.57 & 0.04 & 0.95 & 0.99 & 0.97 & 0.72 & 0.17 & 0.94 & 0.99 & 15.86 \\
  \hline
  
  \multicolumn{16}{|l|}{\textbf{Neural Models}} \\
  \hline
  ProdLDA & 0.60 & -0.02 & 1.00 & 1.00 & 165.04 & 0.66 & 0.05 & 1.00 & 1.00 & 218.76 & 0.39 & -0.19 & 0.89 & 0.98 & 2354.88 \\
  \hline
  Neural LDA & 0.55 & -0.24 & 1.00 & 1.00 & 162.51 & 0.54 & -0.30 & 1.00 & 1.00 & 213.86 & 0.42 & -0.22 & 0.96 & 0.99 & 2186.46 \\
  \hline
  ETM (FastText) & 0.49 & -0.02 & 0.92 & 0.95 & 13.63 & 0.40 & -0.14 & 0.81 & 0.94 & 20.15 & 0.49 & -0.12 & 0.08 & 0.00 & 410.92 \\
  \hline
  Combined TM (BOW+SBERT) & 0.67 & 0.04 & 1.00 & 1.00 & 173.91 & 0.64 & 0.05 & 1.00 & 1.00 & 173.39 & 0.71 & 0.14 & 1.00 & 1.00 & 685.30 \\
  \hline
  ZeroShot TM (BOW+SBERT) & 0.66 & 0.04 & 0.87 & 0.93 & 103.74 & 0.64 & 0.05 & 1.00 & 1.00 & 115.00 & 0.60 & 0.05 & 0.37 & 0.75 & 352.85 \\
  \hline
  
  \multicolumn{16}{|l|}{\textbf{Cluster-Based Models}} \\
  \hline
  Top2vec (Doc2Vec) & 0.83 & 0.18 & 1.00 & 1.00 & 31.98 & 0.74 & 0.11 & 1.00 & 1.00 & 133.64 & 0.80 & 0.12 & 0.99 & 1.00 & 1650.72 \\
  \hline
  BERTopic (Word2Vec+UMAP+HDBSCAN) & 0.26 & -0.15 & 0.78 & 0.81 & 34.78 & 0.44 & -0.05 & 0.73 & 0.90 & 33.58 & 0.39 & -0.12 & 0.85 & 0.95 & 175.85 \\
  \hline
  BERTopic (GloVe+UMAP+HDBSCAN) & 0.53 & 0.03 & 0.83 & 0.88 & 18.70 & 0.56 & 0.13 & 0.68 & 0.91 & 20.40 & 0.43 & -0.11 & 0.83 & 0.95 & 277.12 \\
  \hline
  BERTopic (Doc2Vec+UMAP+HDBSCAN) & 0.25 & -0.16 & 0.67 & 0.72 & 35.96 & 0.44 & -0.05 & 0.72 & 0.90 & 33.57 & 0.36 & -0.12 & 0.84 & 0.94 & 156.38 \\
  \hline
  BERTopic (FastText+UMAP+HDBSCAN) & 0.25 & -0.18 & 0.77 & 0.80 & 46.71 & 0.42 & -0.07 & 0.79 & 0.93 & 42.35 & 0.40 & -0.12 & 0.86 & 0.96 & 166.28 \\
  \hline
  BERTopic (SBERT+UMAP+HDBSCAN) & 0.48 & -0.06 & 0.85 & 0.93 & 67.97 & 0.82 & -0.08 & 0.44 & 0.52 & 55.46 & 0.62 & 0.03 & 0.93 & 0.99 & 355.89 \\
  \hline
  BERTopic (SBERT+PCA+HDBSCAN) & 0.55 & -0.03 & 0.99 & 1.00 & 81.39 & 0.78 & -0.05 & 0.49 & 0.61 & 52.98 & 0.62 & 0.00 & 0.93 & 0.99 & 238.23 \\
  \hline
  BERTopic (SBERT+SVD+HDBSCAN) & 0.66 & 0.01 & 1.00 & 1.00 & 81.35 & 0.78 & -0.04 & 0.51 & 0.62 & 52.74 & 0.56 & 0.03 & 1.00 & 1.00 & 284.71 \\
  \hline
  BERTopic (SBERT+HDBSCAN) & 0.65 & 0.02 & 1.00 & 1.00 & 81.58 & 0.77 & -0.04 & 0.55 & 0.74 & 51.90 & 0.53 & 0.02 & 0.97 & 0.99 & 281.03 \\
  \hline
  BERTopic (SBERT+UMAP+KMeans) & 0.71 & 0.08 & 0.99 & 1.00 & 81.39 & 0.69 & 0.02 & 0.97 & 0.99 & 53.41 & 0.67 & 0.07 & 0.94 & 0.99 & 306.82 \\
  \hline
  BERTopic (SBERT+UMAP+Agglomerative) & 0.54 & -0.09 & 0.93 & 0.96 & 80.99 & 0.58 & -0.09 & 0.86 & 0.96 & 52.93 & 0.65 & 0.00 & 0.97 & 0.99 & 231.10 \\
  \hline
  BERTopic (SBERT+UMAP+DBSCAN) & 0.43 & -0.12 & 0.91 & 0.92 & 81.09 & 0.65 & -0.09 & 0.76 & 0.86 & 51.80 & 0.59 & 0.01 & 0.94 & 0.99 & 232.83 \\
  \hline
  BERTopic (SBERT+UMAP+Spectral) & 0.56 & 0.02 & 0.93 & 0.97 & 71.90 & 0.66 & 0.01 & 0.93 & 0.98 & 54.89 & 0.66 & 0.07 & 0.92 & 0.98 & 353.99 \\
  \hline
  CluWords & 0.59 & 0.01 & 0.98 & 0.99 & 256.55 & 0.52 & -0.06 & 0.90 & 1.00 & 445.42 & - & - & - & - & - \\
  \hline
  
  \multicolumn{16}{|l|}{\textbf{Graph-Based Models}} \\
  \hline
  GNTM & 0.62 & 0.07 & 0.93 & 0.94 & 1599.18 & 0.56 & 0.04 & 0.89 & 0.97 & 3459.53 & - & - & - & - & - \\
  \hline
  GCTM & 0.44 & -0.08 & 1.00 & 1.00 & 2236.28 & 0.73 & 0.05 & 1.00 & 1.00 & 4000.40 & - & - & - & - & - \\
  \hline
  Graph2Topic & 0.73 & 0.15 & 0.90 & 0.80 & 205.53 & 0.59 & 0.08 & 0.69 & 0.68 & 279.51 & - & - & - & - & - \\
  \hline
  
  \multicolumn{16}{|l|}{\textbf{Proposed Model}} \\
  \hline
  \textbf{GHTM} & \textbf{0.90} & \textbf{0.28} & \textbf{1.00} & \textbf{1.00} & 22.16 & \textbf{0.87} & \textbf{0.27} & 0.99 & \textbf{1.00} & 13.57 & \textbf{0.82} & \textbf{0.28} & 0.96 & 0.99 & 231.15 \\
  \hline
  \end{tabular}
  \begin{flushleft}
  The results are averaged across three runs. Notably, the efficacy of all models utilizing SBERT are significantly dependent on the selected embedding model. As previously mentioned in Table~\ref{tab:embeddings_model_summary}, we selected \texttt{shihab17/bangla-sentence-transformer}\cite{uddin2024bangla} for Bengali; however, outcomes may vary based on the model selected.
  \end{flushleft}
  \label{tab:full_results}
  \end{adjustwidth}
  \end{table}

Within the \textbf{cluster-based models}, Top2Vec significantly surpassed its competitors, achieving strong coherence ($\mathbf{C_V}$: 0.74–0.83) and high diversity (TD: 0.99-1.00) scores. However, its runtime scales poorly with dataset size (31.98 seconds for \textit{JamunaNews} to 1650.72 seconds for \textit{BanFakeNews}). While Top2Vec achieved strong results with Doc2Vec embeddings, attempts with other available embedding options, such as \texttt{universal-sentence-encoder-multilingual} and \texttt{distiluse-base-multilingual-cased}, yielded only a single topic across Bengali datasets. Conversely, BERTopic maintained reasonable runtimes across different configurations relative to dataset size, yet its efficacy remained lower than that of Top2Vec. BERTopic's modular architecture enabled us to conduct extensive experimentation with Bengali data across various embedding (Word2Vec, GloVe, FastText, Doc2Vec, SBERT), dimensionality reduction (PCA, SVD, UMAP), and clustering (KMeans, Agglomerative, Spectral, DBSCAN, HDBSCAN) model combinations. Our analysis revealed that the SBERT model paired with UMAP and KMeans ($\mathbf{C_V}$: 0.67–0.71) is preferable for Bengali than the default UMAP+HDBSCAN configuration ($\mathbf{C_V}$: 0.48–0.65) of BERTopic. We further observe from the results in Table~\ref{tab:full_results} that BERTopic performs well with the Bengali SBERT model only, compared to other embedding models. CluWords, another cluster-based approach, demonstrated mediocre performance on \textit{JamunaNews} ($\mathbf{C_V}$: 0.59) and \textit{NCTBText} ($\mathbf{C_V}$: 0.52). Its computationally intensive nature prevented evaluation on the larger \textit{BanFakeNews} dataset, highlighting scalability challenges. Overall, cluster-based models demonstrate strong performance across the model categories, but the comparative analysis reveals a fundamental limitation: they are critically dependent on the choice of embedding model.

Modern \textbf{graph-based models} showed promising capabilities but were computationally expensive. The models required substantial GPU memory for operation, exceeding our computational capabilities. Thus, we could not evaluate them on the larger \textit{BanFakeNews} dataset (48,000+ articles). However, GNTM achieved moderate performance ($\mathbf{C_V}$: 0.56–0.62) on the other two datasets. Interestingly, GCTM excels on \textit{NCTBText} ($\mathbf{C_V}$: 0.73) but underperformed on \textit{JamunaNews} ($\mathbf{C_V}$: 0.44), whereas Graph2Topic exhibited the opposite trend. It performed better on \textit{JamunaNews} ($\mathbf{C_V}$: 0.73) than on \textit{NCTBText} ($\mathbf{C_V}$: 0.59). This corpus-dependent behaviour suggests these models are sensitive to dataset characteristics. Moreover, the graph-based models, being recent innovations, lacked support for any language other than English in their implementation. Bengali datasets require a different vectorization method compared to English, which is why we adapted the vectorization pipelines originally designed for these models. In addition, models in this category required significantly longer training durations (205.53-4000.40 seconds) compared to other models, despite being evaluated only on the smaller datasets. Altogether, the computational and linguistic barriers currently limit their applicability for Bengali topic modeling at scale.

\begin{figure}[ht]
  \centering
  \includegraphics[width=1.1\textwidth, height=!]{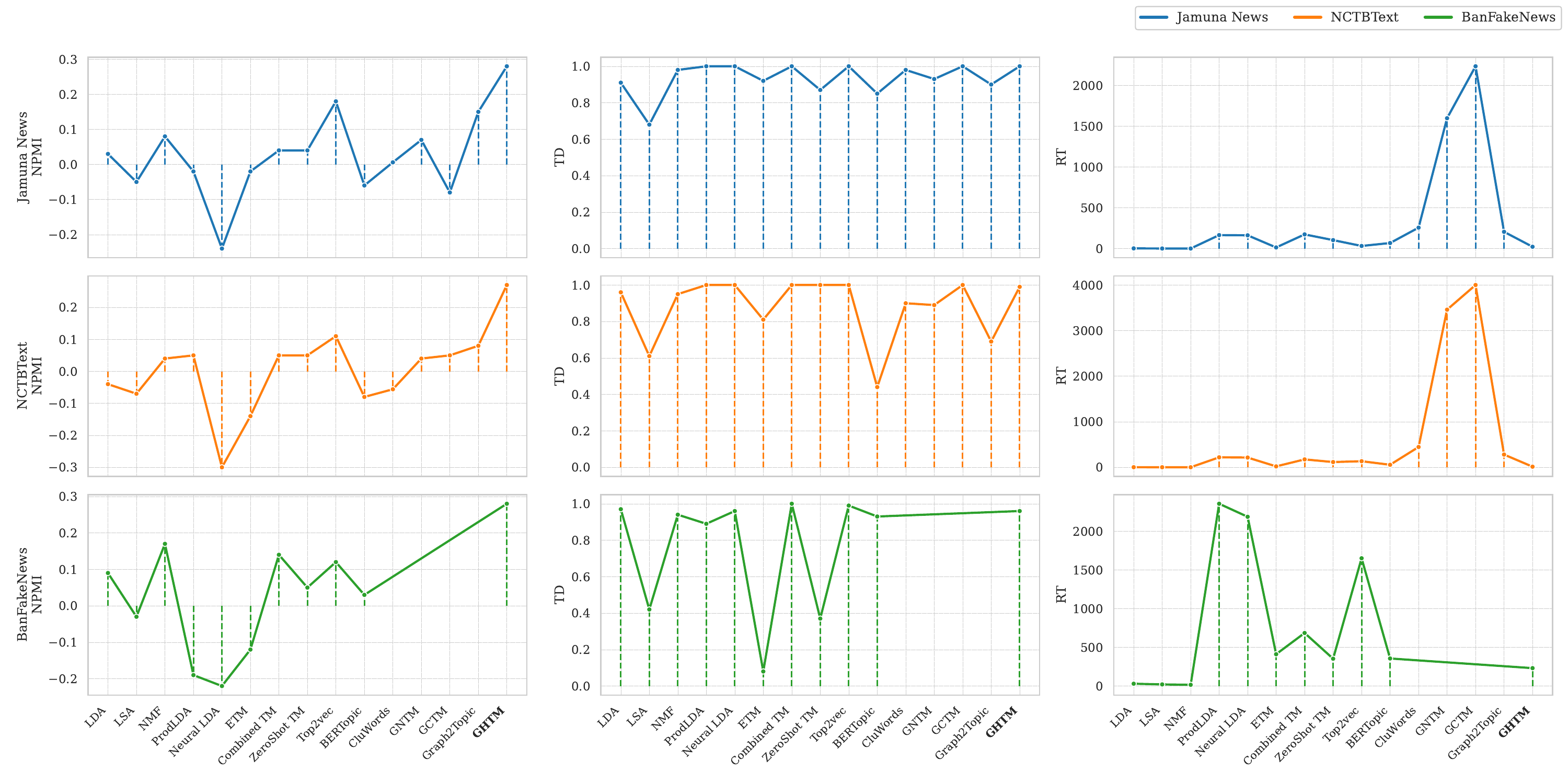}
  \caption{\textbf{Comparative analysis of topic modeling performance across \textit{JamunaNews}, \textit{NCTBText}, and \textit{BanFakeNews} datasets.}
   Each row represents a dataset, with columns illustrating the metrics NPMI, TD and RT respectively. GHTM consistently outperforms baseline models in coherence and diversity. The results of CluWords, GNTM, GCTM, and Graph2Topic for the \textit{BanFakeNews} dataset are absent due to our computational limitations.}
  \label{fig:Fig7}

\end{figure}

\begin{figure}[h!]
  \includegraphics[width=\linewidth]{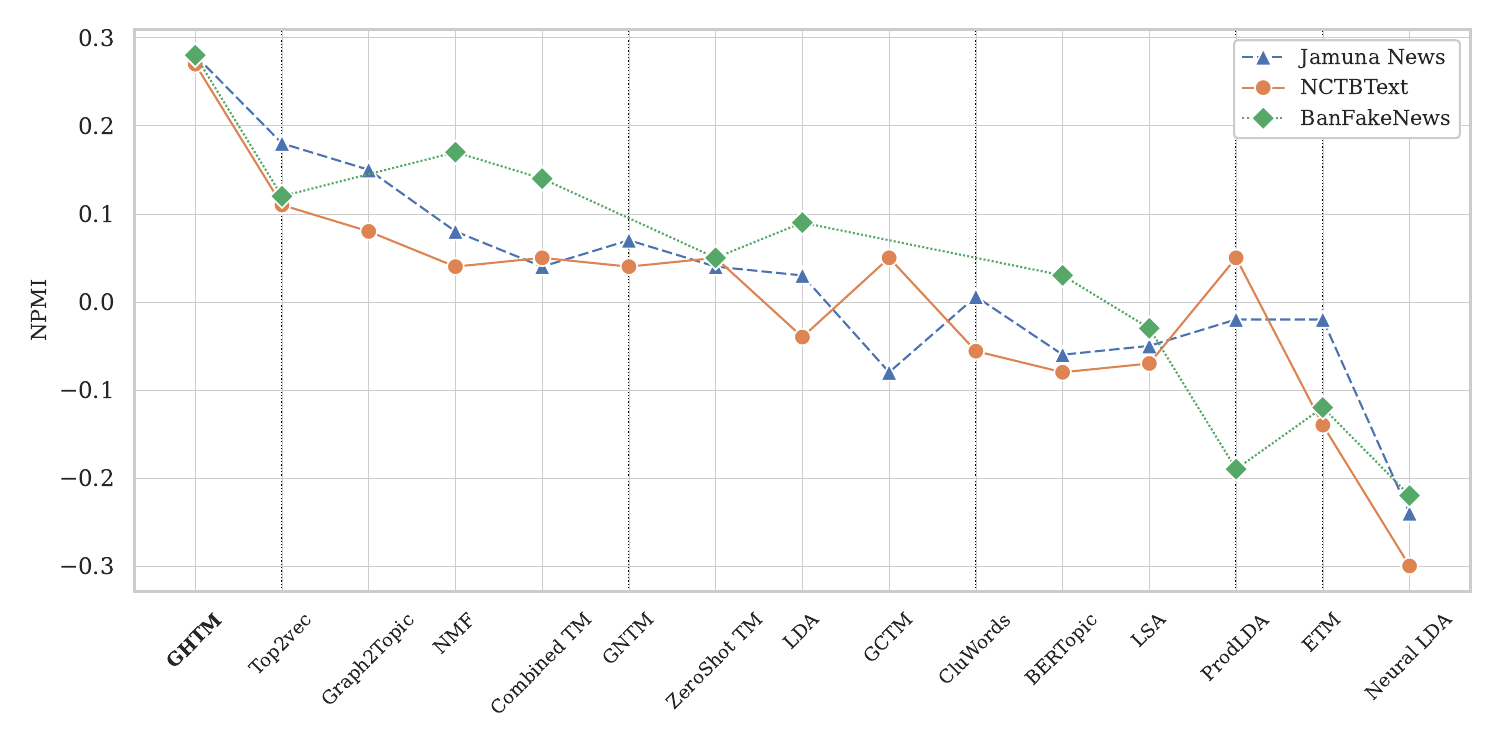}
  \caption{\textbf{NPMI coherence scores across the datasets.} Models are arranged in descending order of their average NPMI scores across the X-axis. Note that graph-based models (Graph2Topic, GNTM, GCTM) and CluWords do not have values for the \textit{BanFakeNews} dataset due to computational constraints on this large dataset.}
  \label{fig:Fig8}
\end{figure}

Among the existing models, Top2Vec achieved overall compelling results, while CombinedTM and Graph2Topic displayed strong performance in their respective categories. However, traditional models like NMF remain competitive for resource-constrained scenarios, offering balanced performance with minimal computational overhead. In contrast, our proposed model, \textbf{GHTM}, demonstrated exceptional performance across all datasets and metrics. Being a mixture of traditional and state-of-the-art components, GHTM achieved substantial improvement over all baseline models, as highlighted in Table~\ref{tab:full_results}. It attained $\mathbf{C_V}$ scores of 0.82–0.90 and NPMI scores of 0.27–0.28 across datasets, representing 8–17\% improvement over the next-best model (Top2Vec: $\mathbf{C_V}$ 0.74–0.83, NPMI 0.11–0.18). It also achieved impressive topic diversity scores (TD: 0.96–1.00, IRBO: 0.99–1.00) while maintaining low runtimes (13.57–231.15 seconds), representing a favorable balance between performance and efficiency. Fig~\ref{fig:Fig7} visualizes GHTM's consistent dominance across datasets and metrics, while Fig~\ref{fig:Fig8} illustrates the substantial coherence gap between GHTM and baseline models. Besides, our experiments revealed that considering bigrams along with unigrams while generating vocabulary substantially boosts the performance of GHTM. However, bigrams were not employed in this study to ensure uniformity in the experimental setup across the models. 

On the \textit{NCTBText} dataset, which comprises non-news sources, topic models show interesting performance variations. Among baseline models, Top2Vec ($\mathbf{C_V}$: 0.74) and GCTM ($\mathbf{C_V}$: 0.73) emerged as strong performers, though at the cost of substantially longer runtimes. Interestingly, several BERTopic configurations showed improved performance on \textit{NCTBText} compared to news datasets. The SBERT+UMAP+HDBSCAN configuration achieved high coherence ($\mathbf{C_V}$: 0.82), notably surpassing its performance on \textit{JamunaNews} ($\mathbf{C_V}$: 0.48) and \textit{BanFakeNews} ($\mathbf{C_V}$: 0.62). Similarly, ProdLDA, which performed poorly on newspaper datasets, demonstrated competitive coherence ($\mathbf{C_V}$: 0.66) on \textit{NCTBText}. Most notably, GHTM maintained consistently high performance across both news and non-news corpora, demonstrating strong generalization capabilities. This consistency, combined with fast runtime on \textit{NCTBText} (13.57s), establishes GHTM as particularly suitable for diverse Bengali text collections beyond traditional news domains.

\subsection*{Performance evaluation of graph-based topic models on English}
To assess the generalizability of GHTM beyond Bengali text, we conducted a comparative analysis against several state-of-the-art graph-based topic models on the English benchmark dataset \textit{20NewsGroup}. We evaluated model performance by averaging the NPMI coherence scores across selected topic numbers, $K \in \{10, 20, 50\}$.

\begin{figure}[h!]
  \includegraphics[width=\linewidth]{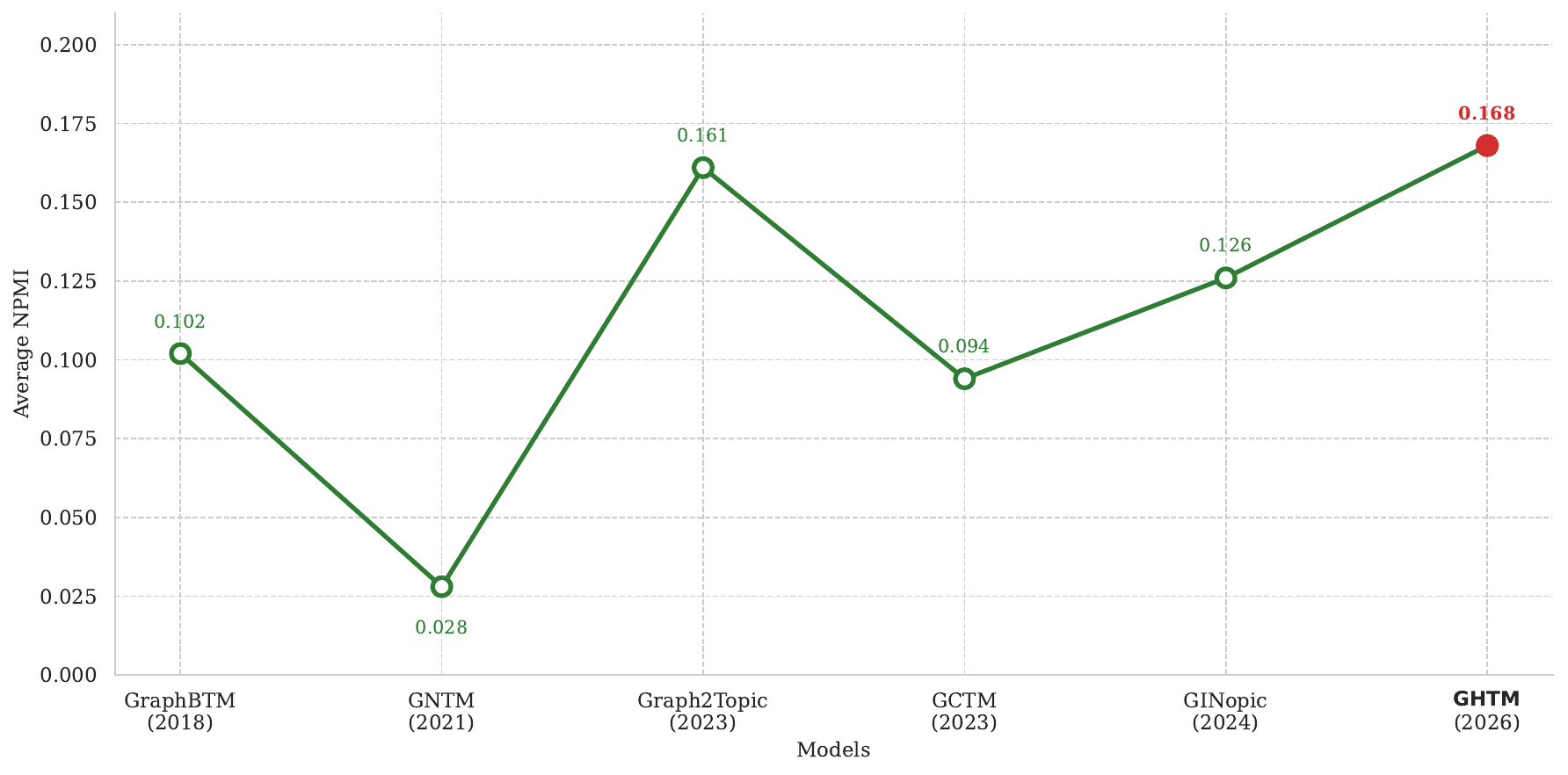}
  \caption{\textbf{Performance Comparison of Graph-Based Topic Models on \textit{20NewsGroup} dataset.} Models are arranged in chronological order by publication year.}
  \label{fig:Fig9}
\end{figure}

As presented in Table~\ref{tab:npm_english}, GHTM demonstrated excellent performance against other graph-based models in English, achieving the highest average NPMI score of 0.168, despite being primarily designed for topic modeling in Bengali. Among the baseline models, GINopic and Graph2Topic exhibited competitive performance, where Graph2Topic achieved an average NPMI of 0.161. However, it is worth noting that Graph2Topic consistently generated an empty list of topic words during evaluation, resulting in the actual number of discovered topics being one less than specified. The comparative performance trends across all evaluated models are visually illustrated in Fig~\ref{fig:Fig9}, highlighting the superiority of GHTM.

\begin{table}[!ht]
  \centering
  \caption{\textbf{Comparison of NPMI scores for graph-centric models on the English benchmark dataset 20NewsGroup}}
  \renewcommand{\arraystretch}{1.5}
  \begin{tabular}{|l|c|c|c|c|}
  \hline
  \textbf{Model} & \textbf{$K=10$} & \textbf{$K=20$} & \textbf{$K=50$} & \textbf{Avg.} \\
  \hline
  GraphBTM & 0.066 & 0.114 & 0.127 & 0.102 \\
  \hline
  GNTM & 0.020 & 0.029 & 0.034 & 0.028 \\
  \hline
  GCTM & 0.098 & 0.053 & \textbf{0.130} & 0.094 \\
  \hline
  GINopic & 0.139 & 0.127 & 0.110 & 0.126 \\
  \hline
  Graph2Topic & 0.183 ($K=9$) & 0.176 ($K=19$) & 0.125 ($K=49$) & 0.161 \\
  \hline
  \textbf{GHTM (Proposed)} & \textbf{0.199} & \textbf{0.189} & 0.115 & \textbf{0.168} \\
  \hline
  \end{tabular}
  \begin{flushleft} 
    NPMI scores for contemporary graph-based topic modeling methods across varying numbers of topics ($K = 10, 20, 50$). \textbf{GHTM} model achieves the highest average NPMI, indicating superior topic coherence. Graph2Topic consistently generated a list of empty strings for one topic during each run, resulting in discovered topics ($K - 1$) that is lower than the specified value ($K$).
  \end{flushleft}
  \label{tab:npm_english}
\end{table}

\subsection*{Topic quality analysis} 
We conducted a qualitative assessment of the topic models to evaluate their performance beyond mathematical criteria. To fairly assess a topic model's ability to generate relevant, coherent, and meaningful words for a given topic, human judgment is required~\cite{Kardam2026}. Also, manual inspection of topic words can reveal out-of-topic or irrelevant words, indicating topic drift or a model's failure to capture the corpus's underlying thematic structure. In this study, the authors serving as domain experts, independently reviewed the top 10 words generated by selected models across specific domains to determine their practical utility.

We do the assessment by carefully examining the topic words generated by the models. Table \ref{tab:topic_words_comparison} displays the top ten topic words in the category Sports, generated by the significant models for the Bengali dataset \textit{JamunaNews}. Except for LDA, BERTopic, GCTM, and Graph2Topic, all models, including GHTM, produced relevant topic words that depicted the sports theme. We know from Table~\ref{tab:dataset_summary} that \textit{JamunaNews} has four classes, or broad topics. Graph2Topic was only able to capture the topics Entertainment, National, and International News, omitting Sports, and producing another redundant set of topic words closely related to Entertainment.
In the case of English, Table~\ref{tab:electronics_topic_words} shows that all models produced a fairly good set of topic words for the Electronics (sci.electronics) category in the \textit{20NewsGroup} dataset, except for Graph2Topic. 

\begin{table}[!ht]
  % \begin{adjustwidth}{-2.0in}{0in} 
  \centering
  \scriptsize
  \caption{\textbf{Top 10 topic words identified by different models in Sports category for Bengali.}}
  \renewcommand{\arraystretch}{1.5}
  \begin{tabular}{|l|l|l|c|}
    \hline
    \textbf{Model} & \textbf{Top 10 Topic Words} & \textbf{Out-of-Topic Words} & \textbf{Count} \\
    \hline
    LDA & \bn{রান, ম্যাচ, বাংলাদেশ, দল, বিশ্বকাপ, নির্বাচন, ভারত, ক্রিকেট, উইকেট, সিরিজ} & \bn{নির্বাচন} & 1 \\
    \hline
    LSA & \bn{রান, গোল, ম্যাচ, উইকেট, মেসি, খেলা, ফাইনাল, ফুটবল, আর্জেন্টিনা, লিগ} & - & 0 \\
    \hline
    NMF & \bn{রান, ম্যাচ, বিশ্বকাপ, বাংলাদেশ, দল, উইকেট, খেলা, হারা, জয়, টোয়েন্ট} & - & 0 \\
    \hline
    Combined TM & \bn{হারা, বিশ্বকাপ, টোয়েন্টি, উইকেট, ম্যাচ, রান, দল, খেলা, বাংলাদেশ, অধিনায়ক} & - & 0 \\
    \hline
    ZeroShot TM & \bn{রান, ম্যাচ, বাংলাদেশ, দল, উইকেট, বিশ্বকাপ, হারা, জয়, খেলা, পক্ষ} & - & 0 \\
    \hline
    Top2Vec & \bn{ম্যাচে, ম্যাচ, বিপক্ষে, অধিনায়ক, ম্যাচের, রান, উইকেট, বল, ব্যাট, জয়ের} & - & 0 \\
    \hline
    BERTopic & \bn{বিশ্বকাপের, ভারত, বিশ্বকাপ, অভিনেত্রী, ম্যাচ, বলিউড, ভারতের, ম্যাচে, সিনেমা, অভিনেতা} & \bn{অভিনেত্রী, বলিউড, সিনেমা, অভিনেতা} & 4 \\
    \hline
    CluWords & \bn{ক্রিকেট, ক্রিকেটে, ম্যাচে, ম্যাচ, বিশ্বকাপটা, বিশ্বকাপে, ম্যাচের, বিশ্বকাপের, বিশ্বকাপ, খেলার} & - & 0 \\
    \hline
    GNTM & \bn{বাংলাদেশ, সংগৃহীত, বিপক্ষে, ম্যাচে, ম্যাচ, দলের, ম্যাচের, রান, বিশ্বকাপ, মাঠে} & - & 0 \\
    \hline
    GCTM & \bn{টিভিতে, চলুন, সময়সূচি, স্বাগতিকরা, টেবিলের, ব্যবধান, অলআউট, মিনিটেই, প্রোটিয়াদের, যাক} & \bn{টিভিতে, চলুন, সময়সূচি, স্বাগতিকরা, টেবিলের, মিনিটেই, যাক} & 7 \\
    \hline
    Graph2Topic & \bn{।, বিয়ে, বিয়ের, , , আমার, তার, অভিনেত্রী, নিয়ে, বিচ্ছেদের, তিনি} & \bn{।, বিয়ে, বিয়ের, , , আমার, তার, অভিনেত্রী, নিয়ে, বিচ্ছেদের, তিনি} & 10 \\
    \hline
    \textbf{GHTM (Proposed)} & \bn{রান, উইকেট, ইনিংস, ব্যাট, নামা, শামি, সেঞ্চুরি, কোহলি, অস্ট্রেলিয়া, পাঞ্জাব} & - & 0 \\
    \hline
  \end{tabular}
  \begin{flushleft} 
    Qualitative analysis of the top 10 topic words identified using topic modeling approaches in the Bengali Sports category for \textbf{JamunaNews} dataset. The table displays the most representative words extracted by each model, as well as manually identified out-of-topic words that are not semantically related to the sports domain. The fourth column shows the count of irrelevant words in the top 10 words. The proposed GHTM model successfully generates domain-relevant topic words while producing zero out-of-topic terms, demonstrating superior topic quality. 
  \end{flushleft}
  \label{tab:topic_words_comparison}
  % \end{adjustwidth}
  \end{table}
  
\begin{table}[!ht]
  % \begin{adjustwidth}{-2.0in}{0in}
  \centering
  \scriptsize
  \caption{\textbf{Top 10 topic words identified by different models in the Electronics category for English}}
  \renewcommand{\arraystretch}{1.5}
  \begin{tabular}{|l|l|l|c|}
    \hline
    \textbf{Model} & \textbf{Top 10 Topic Words} & \textbf{Out-of-Topic Words} & \textbf{Count} \\
    \hline
    GraphBTM & scsi, isa, ide, bus, scsus, mb, ram, jumper, pc, dos & - & 0 \\
    \hline
    GNTM & drive, disk, controller, ide, hard, drives, bus, card, scsi, use & use & 1 \\
    \hline
    GCTM & circuit, switch, port, cable, serial, input, signal, output, floppy, drive & - & 0 \\
    \hline
    GINopic & drive, monitor, card, software, port, cable, system, modem, disk, pin & - & 0 \\
    \hline
    Graph2Topic & ibm, controller, beamer, yang, bootbzip, computer, thank, adressing, terminal, compaq & thank, yang & 2 \\
    \hline
    \textbf{GHTM (Proposed)} & scsi, mb, drive, controller, chip, meg, mbs, ide, simm, socket & - & 0 \\
    \hline
  \end{tabular}
  \begin{flushleft}
  This table provides a qualitative comparison of topic modeling performance in the Electronics category for the English dataset \textbf{20NewsGroup}. Out-of-topic words are terms that have no semantic relevance to electronics or hardware components. All models successfully capture computer hardware-specific terminology, with the exception of Graph2Topic, which includes conversational terms (``thank") and potentially noisy tokens (``yang"). The proposed GHTM model exhibits strong domain coherence by identifying only technical terms related to storage devices, memory components, and computer architecture with no semantic drift.
  \end{flushleft}
  \label{tab:electronics_topic_words}
  % \end{adjustwidth}
  \end{table}
  
\subsection*{Ablation study} 
To better understand the role of individual components in GHTM, we perform a comprehensive ablation study that involves systematically removing and replacing key architectural elements. This analysis provides insights into the design choices that drive the model's performance and validates the importance of each component.

Table~\ref{tab:ablation} shows the NPMI scores for different component combinations based on the \textit{JamunaNews} dataset. Our analysis reveals that, TF-IDF vectorization outperforms BoW representation, indicating that a term weighting scheme provides more discriminative features for topic extraction. Moreover, among the clustering and matrix factorization approaches that we tested, standard Non-negative Matrix Factorization (NMF) demonstrates superior performance compared to KMeans clustering, Spectral Clustering, and Semi NMF~\cite{ding2010convex}.
\begin{table}[!ht]
  \centering
  \caption{\textbf{Ablation study results.}}
  \begin{tabular}{|l|c|}
  \hline
  \textbf{Variant} & \textbf{NPMI} \\
  \hline
  TF-IDF + NMF & 0.08 \\
  \hline
  TF-IDF * GloVe + NMF & 0.18 \\
  \hline
  TF-IDF * GloVe + GCN + KMeans & 0.16 \\
  \hline
  TF-IDF * GloVe + GCN + Spectral & 0.22 \\
  \hline
  \textbf{TF-IDF * GloVe + GCN + NMF (GHTM)} & \textbf{0.28} \\
  \hline
  \end{tabular}
  \label{tab:ablation}
  \end{table}

The comparison of NMF and Semi-NMF is especially noteworthy. Semi-NMF ~\cite{ding2010convex} relaxes the non-negativity constraint on the input matrix while maintaining non-negative factors in the output, making it theoretically more suitable for graph embeddings with potentially negative values. However, our findings show that using the absolute value function to convert GCN-generated embeddings into non-negative representations prior to standard NMF does not degrade performance. In fact, this approach produces slightly higher topic coherence scores than Semi-NMF. This implies that, while the sign information in graph embeddings is lost during transformation, the magnitude and relative ordering of embedding dimensions provide enough discriminative information for effective topic modeling. 

\subsection*{Evolution of the embedding space}
\begin{figure}[h!]
  \includegraphics[width=\linewidth]{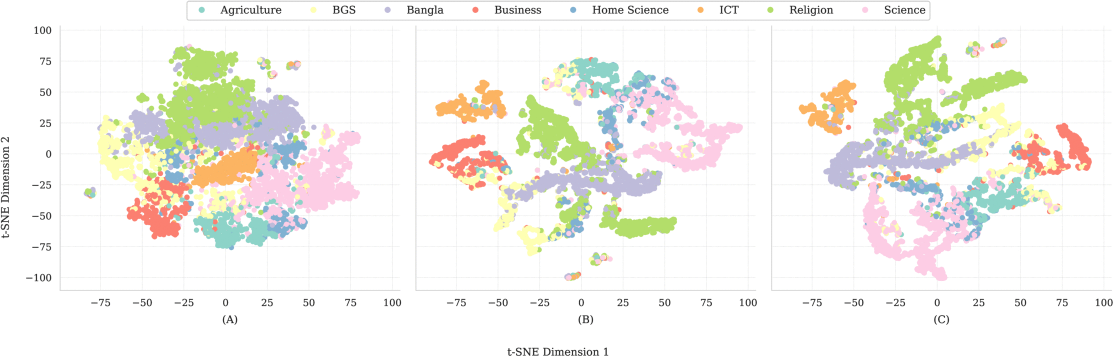}
  \caption{\textbf{Embedding space evolution and cluster separability across transformation stages.} Two-dimensional t-SNE projection of document representations from the NCTBText dataset, at three sequential processing stages. (A) Following text vectorization (TF-IDF × GloVe). (B) Following Cluster-GCN propagation. (C) Following the absolute value transformation, applied before matrix factorization. Axes in all panels denote t-SNE dimensions.}
  \label{fig:Fig10}
\end{figure}

To qualitatively evaluate the progressive refinement of the document representations, we employ t-SNE projections at three critical stages of GHTM. Fig~\ref{fig:Fig10} illustrates how each processing stage reshapes the geometric organization of the embedding space on the NCTBText dataset.

Fig~\ref{fig:Fig10}A illustrates the initial vector space resulting from the text vectorization ($\text{TF-IDF} \times \text{GloVe}$) stage, where topic-related structures remain diffuse and loosely defined. Subsequently, Fig~\ref{fig:Fig10}B shows the post-propagation space following the second stage (graph-based representation learning). As this stage aggregated neighborhood signals over the document graph, semantically similar documents converged into compact groupings. As a result, inter-class margins widened, producing markedly more cohesive and well-separated clusters relative to Fig~\ref{fig:Fig10}A. Finally, Fig~\ref{fig:Fig10}C depicts the non-negative embedding space produced by the absolute value transformation, applied prior to matrix factorization in the third stage. Although this operation projects the representations into a bounded positive subspace, it preserves the cluster separability established in Fig.\ref{fig:Fig10}B without structural degradation. Together, these transformation stages yield a non-negative feature space with semantically well-separated regions, effectively preparing the space for downstream topic extraction via matrix factorization.

\subsection*{Statistical significance test} 
To determine if the observed differences in topic coherence between models for Bengali, were statistically significant, we used the non-parametric Friedman test, followed by a Nemenyi post-hoc analysis. This framework is a recommended method for comparing multiple algorithms across different datasets\cite{Demsar2006}.

The Friedman test ranks the models for each dataset (1 for the best performer, 2 for the second, and so on) and averages their rankings across all datasets. It tests the null hypothesis, which states that all models perform equally. A rejected hypothesis means that at least one model is significantly different. Our test used a matrix of NPMI coherence scores (averaged over three runs) for the eleven non-graph-based models across the three datasets (\textit{JamunaNews}, \textit{NCTBText}, and \textit{BanFakeNews}). Graph-based models were excluded from this analysis because we do not have scores for the \textit{BanFakeNews} dataset. The test yielded a test statistic of $\chi^2 =$ 7.476 and a p-value of 0.0238, which is less than our significance level of $\alpha = 0.05$. This allows us to reject the null hypothesis and conclude that there are statistically significant differences between the model performances.

\begin{figure}[h!]
  \includegraphics[width=\linewidth]{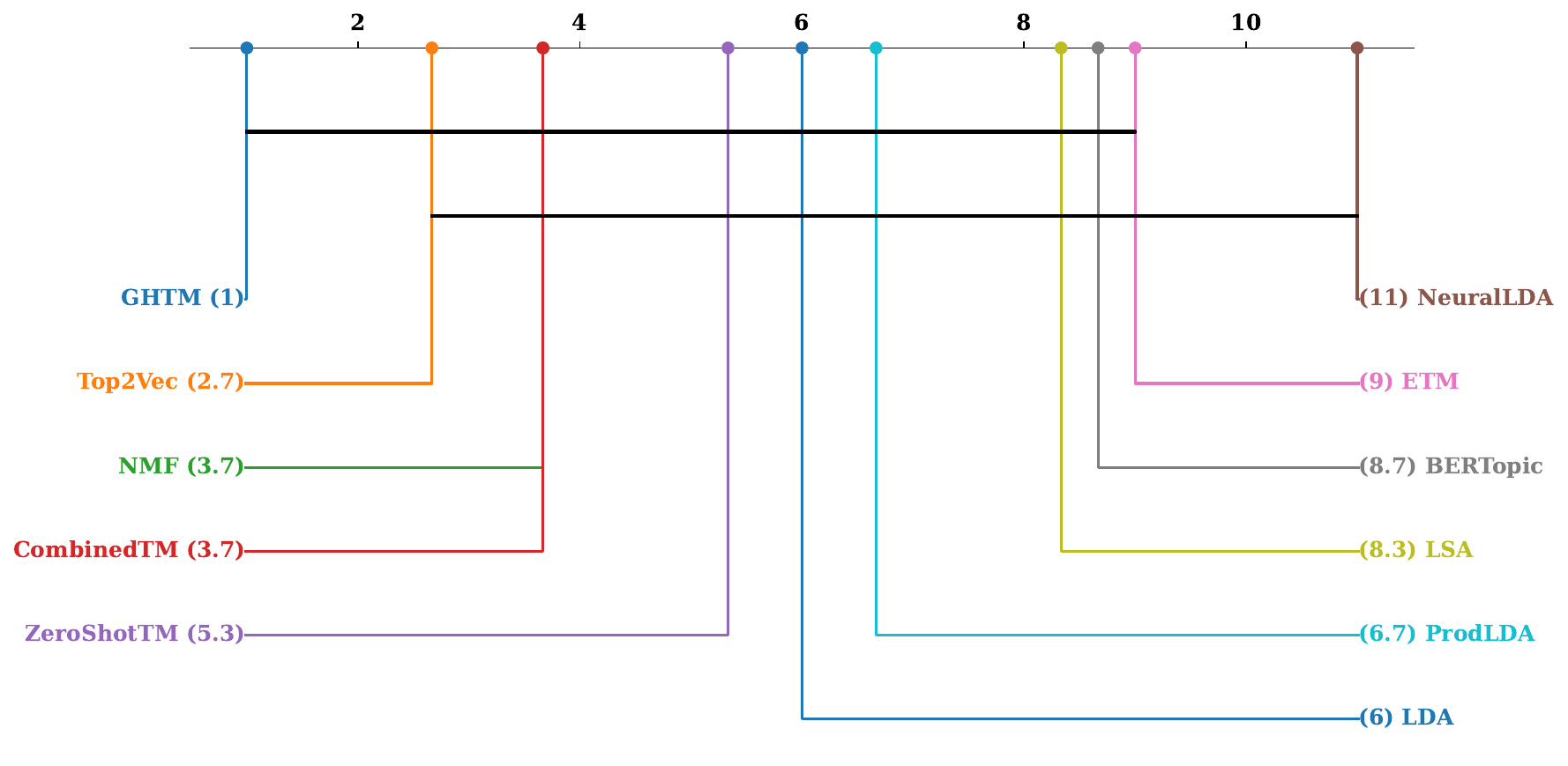}
  \caption{\textbf{Critical Difference Diagram.} Visualizes the Nemenyi post-hoc test results. The proposed GHTM model (far left) obtained the best average rank.}
  \label{fig:Fig11}
\end{figure}

To determine the statistical significance in difference between GHTM and other baselines, we used the Nemenyi post-hoc test. This test adjusts the p-values for all pairwise comparisons to reduce the family-wise error rate, making it more conservative and robust than multiple individual tests. The results of this test are summarized in Table~\ref{tab:average_ranks}, which lists the models ordered by their average rank (1.0 represents the best possible rank). Our proposed model, GHTM, received the highest ranking. The final column shows the Nemenyi-adjusted p-values for each model versus GHTM. The analysis shows that GHTM's performance is significantly better than only NeuralLDA. Although GHTM achieved a higher average rank than other baselines, the Nemenyi test did not deem these differences statistically significant at the $\alpha = 0.05$ level. This result is likely influenced by the limited number of datasets ($n=3$) in our study and the conservative nature of the Nemenyi test. However, the substantial difference in average rank between GHTM and the top baselines suggests a performance advantage, that may be confirmed with further validation on more datasets. The critical difference diagram\cite{Demsar2006} in Fig~\ref{fig:Fig11} further illustrates these findings, showing that GHTM, positioned at the far left, achieved the best average rank.

\begin{table}[!ht]
  \centering
  \small
  \caption{\textbf{Nemenyi post-hoc test results showing average ranks and adjusted p-values of baseline models compared to GHTM.}}
  \renewcommand{\arraystretch}{1.3}
  \begin{tabular}{|l|c|c|c|c|c|}
  \hline
  \textbf{Model} & \textbf{Avg. Rank} & \textbf{Adj. p-value} \\
   &  & \textbf{(vs. GHTM)} \\
  \hline
  GHTM & 1.0 & -- \\
  \hline
  Top2Vec & 2.7 & 0.999 \\
  \hline
  NMF & 3.7 & 0.996 \\
  \hline
  CombinedTM & 3.7 & 0.994 \\
  \hline
  ZeroShotTM & 5.3 & 0.945 \\
  \hline
  LDA & 6.0 & 0.752 \\
  \hline
  ProdLDA & 6.7 & 0.450 \\
  \hline
  LSA & 8.3 & 0.196 \\
  \hline
  BERTopic & 8.7 & 0.146 \\
  \hline
  ETM & 9.0 & 0.125 \\
  \hline
  NeuralLDA & 11.0 & 0.010 \\
  \hline
  \end{tabular}
  \label{tab:average_ranks}
\end{table}

\section*{Discussion}
This study advances Bengali topic modeling by addressing four research questions. We sought to inquire (1) which topic models from traditional to contemporary perform best for Bengali, (2) whether a novel hybrid model combining top-performing strategies could yield superior results, (3) how such a novel model would perform on English, and (4) how both the new model and existing approaches would perform on a novel Bengali corpus curated from non-newspaper sources. Through this inquisition, our research establishes a comprehensive benchmark for evaluating topic models in Bengali, introduces a novel topic model, and validates its effectiveness across Bengali and English. It also presents a novel dataset, proposed as a benchmark for Bengali topic modeling, which significantly broadens the lexical resources within Bengali NLP corpora. In this section, we discuss our key contributions, findings, and their implications through the lens of our research questions and elaborate on the limitations and future directions of this work.

To address our first research question, we conducted a thorough evaluation across three different datasets, using five standard metrics and a diverse set of baseline models. The comprehensive evaluation offers the first systematic account of how models spanning from conventional to contemporary perform on Bengali data and, in doing so, marks a pivotal shift in Bengali topic modeling research. Prior studies\cite{helal2018topic,hasan2019lda2vec,alam2020bengali,paul2025combining,rifat2025clustering,dawn2024likelihood} lacked comparative analysis across model categories, standardized evaluation frameworks, and primarily used LDA as a common baseline without considering contemporary methodologies, which made objective assessment very difficult. By systematically evaluating traditional and current models across diverse datasets, our work fills this long-standing gap. Therefore, it establishes the first rigorous benchmark in the Bengali topic modeling literature, reported in Table~\ref{tab:full_results}.

Besides, the benchmark simultaneously highlights the key limitations of existing approaches. Notably, neural topic models and, more critically, graph-based models, despite their architectural sophistication, faced severe scalability constraints due to substantial GPU memory requirements and high training times even on smaller datasets. Although cluster-based models showed competitive performance overall, their effectiveness remains heavily dependent on embedding model selection, for which pre-trained models in Bengali remain scarce.

Moreover, the benchmarking framework not only provides a foundation for reproducible evaluation but also offers a starting point for future researchers in this field, enabling meaningful comparisons and incremental progress. Most importantly, it revealed a critical void in the landscape: the field lacked a model that simultaneously achieves computational efficiency, scalability to large datasets, and reduced dependency on embedding models; a combination essential for practical Bengali NLP applications. This realization directly motivated the development of our hybrid topic model, as we discuss in our response to the second research question below.

To answer our second research question, we leveraged the insights derived from the benchmark results to develop GHTM. In particular, we observed the effectiveness of GloVe embeddings compared to other word-level vectorization approaches. We also observed NMF, a simple algebraic method, being remarkably efficient as a standalone topic modeling technique, despite the advent of more complex neural architectures. Building on these insights, we designed GHTM's architecture to combine complementary strengths of multiple paradigms. The model combines statistical term weighting (TF-IDF) with GloVe's semantic word embeddings, incorporates GCN to enrich these representations, and utilizes NMF to factorize the enriched embeddings into interpretable topics. This strategic combination of powerful components resulted in a model that directly addresses the key limitations of existing approaches. Specifically, GHTM avoids the computational intensity of pure neural and graph-based models by using efficient matrix factorization for topic extraction, maintains scalability through strategic use of ClusterGCN, and reduces embedding dependency by combining multiple representation strategies. As a result, GHTM successfully outperforms all baseline models in topic coherence and diversity across all Bengali datasets while maintaining minimal runtime, as evidenced by Table~\ref{tab:full_results} and further illustrated in Fig~\ref{fig:Fig7} and Fig~\ref{fig:Fig8}. Moreover, as shown in Table~\ref{tab:topic_words_comparison}, GHTM generates topic words that are not only quantitatively outstanding but also semantically coherent and meaningful. Therefore, the success of GHTM validates our hypothesis that a carefully designed hybrid approach can surpass individual methodologies. 

However, the implications of GHTM extend beyond immediate performance gains. GHTM represents a significant methodological departure from prior Bengali topic modeling research, which mostly explored LDA\cite{helal2018topic,alam2020bengali} and its variants\cite{hasan2019lda2vec,paul2025combining}. Though, a few recent studies have introduced alternative approaches, such as LCD\cite{dawn2024likelihood} and a cluster-based method\cite{rifat2025clustering}, but they remain close to conventional paradigms and do not leverage the representational benefits of modern techniques. GHTM, on the other hand, introduces a hybrid methodology that blends conventional techniques (TF-IDF, NMF) with modern methods (GloVe, GCN). The model unifies statistical (TF-IDF), semantic (GloVe), and structural (GCN) representations, demonstrating that synthesizing complementary paradigms across different methodological traditions can yield stronger performance and greater practical viability. 

GHTM's high-quality unsupervised topic modeling, proven by consistently excellent performance, has significant potential to advance Bengali NLP by facilitating multiple downstream tasks, including unsupervised document classification, information retrieval, exploratory data analysis, and semi-supervised data annotation workflows. For low-resource languages like Bengali, where labeled datasets remain scarce and expensive to produce, we can particularly benefit from automated data annotation. Besides, GHTM's success validates a crucial principle: a thoughtful combination of complementary techniques informed by systematic evaluation can be more effective than pursuing increasingly complex architectures. This insight has broader implications for NLP research in low-resource contexts, suggesting that the strengths of different methodological paradigms can produce innovations that exceed the sum of their parts.

In response to our third research question, we evaluated GHTM on \textit{20Newsgroup}, a widely recognized English benchmark dataset for topic modeling research. As Table~\ref{tab:npm_english} shows, GHTM generalizes effectively to English, achieving excellent performance without any language-specific adaptation. Despite being primarily designed for Bengali, GHTM outperformed well-established graph-based models for English, including GINopic, Graph2Topic, GNTM, GCTM, and GraphBTM. By providing practical value beyond Bengali alone, it exemplifies that efficient hybrid designs can transcend linguistic boundaries. Moreover, the quality of the topics discovered by GHTM for \textit{20Newsgroup}, as shown in Table~\ref{tab:electronics_topic_words}, further validates the model's ability to correctly identify the latent topical structure in English. In other words, the findings conclusively address our third research question, confirming that GHTM's effectiveness extends beyond Bengali and establishing its viability as a language-agnostic topic modeling approach suitable for diverse multilingual applications.

Finally, to answer the fourth research question, we curated the \textit{NCTBText} dataset from educational textbooks to examine how models perform on a fresh non-newspaper corpus. Evaluating models on diverse datasets is crucial for developing robust models that generalize across domains and different text types, revealing performance characteristics that would remain hidden in news-only evaluations. Table~\ref{tab:full_results} demonstrates that GHTM achieves the best performance on \textit{NCTBText} among all evaluated models. Interestingly, several baseline models, such as BERTopic and ProdLDA, exhibited notably higher performance on \textit{NCTBText} compared to newspaper datasets. While BERTopic have demonstrated practical value and interpretability in domain mapping over large-scale, high-resource research corpora~\cite{gupta2024generative} due to their data-intensive nature, our findings reveal that in low-resource linguistic settings, their performance becomes corpus-dependent and sensitive to domain variations. This further emphasizes the necessity of benchmarking models across diverse text types to determine their true practical viability. The observations confirm that \textit{NCTBText} successfully serves its intended purpose by revealing model strengths and weaknesses across different corpus characteristics. Besides, the dataset, which spans eight diverse subject areas, provides a linguistically rich alternative to the news-centric datasets that dominate Bengali NLP. It introduces new vocabulary, diverse themes, and varied linguistic structures that differ substantially from journalistic writing and remain largely absent from existing resources. Therefore, the dataset's public availability with labels makes it readily applicable to a range of NLP tasks beyond topic modeling. Moreover, prior Bengali topic modeling studies have lacked a standardized benchmark comparable to the widely used \textit{20Newsgroup} dataset in English. Each study used a different dataset, which hinders meaningful cross-study comparisons. \textit{NCTBText} addresses this gap and is proposed as a pioneering standardized benchmark for evaluating Bengali topic models, supported by its demonstrated lexical diversity and well-separable class structure.

To summarize, this study not only establishes a much-needed benchmark for Bengali topic modeling but also introduces GHTM, which redefines the state-of-the-art in this field. It also establishes \textit{NCTBText}, a gold-standard dataset against which future Bengali topic models can be measured. The contributions provide a solid foundation for future research in Bengali topic modeling and offer insights applicable to other low-resource languages facing similar challenges.

\subsection*{Limitations and future scope}
Despite promising results, this study has a few limitations that point toward directions for future research. First, our model's performance is partly constrained by the availability of Bengali NLP resources. Better foundational tools, such as a GloVe model trained on larger, more diverse pre-training corpora and a more accurate Bengali lemmatizer, could further improve model performance. Second, while our novel \textit{NCTBText} dataset makes an important contribution, it is constrained by the accuracy of available Bengali OCR tools. More precise OCR tools would improve the quality of the dataset. In the future, improvements in these fundamental Bengali NLP tools would benefit not only topic modeling but the entire Bengali NLP ecosystem. Another potential limitation concerns our choice of the sentence embedding model for baseline evaluations: Bengali SBERT. We selected the model that represents the current best option for Bengali, but the selection of a better or worse embedding model could alter comparative results. This highlights the importance of continually re-evaluating benchmarks as Bengali NLP resources improve. 

Looking forward, these limitations suggest several promising research directions. Beyond addressing the specific constraints identified above, a valuable future scope would be adapting GHTM for practical Bengali dataset annotation workflows with relatively minor modifications. By implementing interactive topic merging capabilities, refining document-to-topic assignments, and developing user-friendly interfaces, GHTM could serve as the foundation for an effective semi-supervised annotation tool. Another promising future direction would be to use GHTM-identified top representative documents for a topic as context for an LLM to generate richer, more interpretable topic labels and descriptions. This integration could significantly enhance human understanding of the latent topics of a dataset, extending GHTM's capabilities for exploratory data analysis.  Additionally, adapting mechanisms from related graph methods, such as density-adaptive edge weighting~\cite{AGP2025}, tight-fitting graph construction~\cite{TFGIN2025}, and memory-efficient graph compression~\cite{TianheGraph2025}, could further improve GHTM's overall performance by strengthening its partitioning quality, graph structure, and scalability. These directions collectively point toward a research agenda that builds upon GHTM's foundations to create increasingly practical, accessible, and powerful tools for Bengali NLP.

\section*{Conclusion} 
In this study, we present the first comprehensive evaluation of topic modeling approaches for Bengali, systematically comparing traditional models (LDA, LSA, NMF), neural architectures (ProdLDA, NeuralLDA ETM, CombinedTM, ZeroShotTM), cluster-based methods (Top2Vec, BERTopic, CluWords), and graph-based frameworks (GNTM, GCTM, Graph2Topic) across multiple datasets using standardized metrics. This systematic evaluation not only lays the foundation for future comparative analysis in Bengali topic modeling but also reveals that the field lacked a model that simultaneously achieved computational efficiency, scalability, and robust performance. Therefore, we introduce GHTM, a novel architecture that strategically integrates TF-IDF-weighted GloVe embeddings, GCNs, and NMF. GHTM constructs a document-similarity graph from initial hybrid embeddings, leverages GCN to enhance these representations through iterative neighborhood aggregation, and applies NMF to decompose the graph-enriched embeddings into interpretable topics. The model achieves substantial improvements in topic coherence (NPMI: 0.27-0.28) compared to existing methods, while maintaining computational efficiency with runtimes of 13.57-231.15 seconds across datasets of varying scales. More importantly, GHTM also demonstrates strong performance against several graph-based models such as GINopic, Graph2Topic, GNTM, GCTM, and GraphBTM on the English \textit{20Newsgroups} benchmark. We also introduce \textit{NCTBText}, a diverse Bengali textbook-based dataset curated from eight subject areas. This dataset provides much-needed topical and lexical diversity beyond the newspaper-dominated landscape of Bengali corpora and stands as a pioneering benchmark for Bengali topic modeling. The publicly available implementation of GHTM and the \textit{NCTBText} dataset address the reproducibility challenges that have historically hindered progress in this field. Collectively, these contributions establish a solid foundation for future research, sustained progress in Bengali topic modeling, and can further aid other low-resource languages facing similar challenges.

% \section*{Acknowledgments}
% This work is supported by a research grant (2024-25) from the University of Dhaka funded by the University Grants Commission of Bangladesh. 

\bibliography{ref}
\end{document}